\documentclass{article}
\usepackage{arxiv}

\usepackage[utf8]{inputenc} 
\usepackage[T1]{fontenc}    
\usepackage{lmodern}
\usepackage{hyperref}       
\usepackage{url}            
\usepackage{booktabs}       
\usepackage{amsfonts}       
\usepackage{nicefrac}       
\usepackage{microtype}      
\usepackage{lipsum}		
\usepackage{graphicx}
\usepackage[square,sort,comma,numbers]{natbib}
\usepackage{doi}
\usepackage[dvipsnames, table]{xcolor}%
\usepackage{soul}
\usepackage{subcaption}
\usepackage{physics}
\usepackage{amssymb}
\usepackage{siunitx}
\usepackage{tabularx}
\usepackage{multirow}

\graphicspath{{./images/}}

\newcommand{\coloredhline}[1]{\arrayrulecolor{gray!40}\hline\arrayrulecolor{black}}

\newcommand{\bs}[1]{\vb*{#1}} 

\title{A machine learning framework for uncovering stochastic nonlinear dynamics from noisy data}

\author{
\href{https://orcid.org/0009-0003-5140-7214}{\includegraphics[scale=0.06]{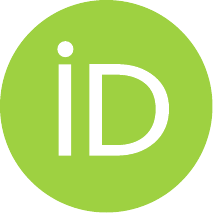}\hspace{1mm}\textcolor{black}{Matteo Bosso}}
\And
\href{https://orcid.org/0000-0002-9863-0291}{\includegraphics[scale=0.06]{orcid.pdf}\hspace{1mm}\textcolor{black}{Giovanni Franzese}}
\And
\href{https://orcid.org/0009-0008-8784-3583}{\includegraphics[scale=0.06]{orcid.pdf}\hspace{1mm}\textcolor{black}{Kushal Swamy}}
\And
\href{https://orcid.org/0000-0002-1545-844X}{\includegraphics[scale=0.06]{orcid.pdf}\hspace{1mm}\textcolor{black}{Maarten Theulings}}
\And
 \href{https://orcid.org/0000-0003-2275-6207}{\includegraphics[scale=0.06]{orcid.pdf}\hspace{1mm}\textcolor{black}{Alejandro M.~Aragón}}\thanks{Corresponding authors. E-mail addresses: \texttt{\href{mailto:a.m.aragon@tudelft.nl}{a.m.aragon@tudelft.nl}} (A.M.~Aragón), \texttt{\href{mailto:f.alijani@tudelft.nl}{f.alijani@tudelft.nl}} (F.~Alijani)}
    \And
	\href{https://orcid.org/0009-0001-5955-8856}{\includegraphics[scale=0.06]{orcid.pdf}\hspace{1mm}\textcolor{black}{Farbod Alijani$^\ast$}} 
	 \AND
	 \\[-1em] 
	 Faculty of Mechanical Engineering, Delft University of Technology\\
	 Mekelweg 2, 2628 CD, Delft, Zuid-Holland,
	 The Netherlands\\	
}

\renewcommand{\shorttitle}{Learning Stochastic Nonlinear Dynamics from Noisy Data}

\hypersetup{
pdftitle={A template for the arxiv style},
pdfsubject={q-bio.NC, q-bio.QM},
pdfauthor={David S.~Hippocampus, Elias D.~Striatum},
pdfkeywords={First keyword, Second keyword, More},
}

\hypersetup{
  colorlinks=true,
  breaklinks=true,
  urlcolor=RoyalBlue,
  linkcolor=RoyalBlue,
  citecolor=RoyalBlue
} 

\begin{document}
\maketitle

\begin{abstract}

Modeling real-world systems requires accounting for noise---whether it arises from unpredictable fluctuations in financial markets, irregular rhythms in biological systems, or environmental variability in ecosystems. While the behavior of such systems can often be described by stochastic differential equations, a central challenge is understanding how noise influences the inference of system parameters and dynamics from data. Traditional symbolic regression methods can uncover governing equations but typically ignore uncertainty. Conversely, Gaussian processes provide principled uncertainty quantification but offer little insight into the underlying dynamics. In this work, we bridge this gap with a hybrid symbolic regression–probabilistic machine learning framework that recovers the symbolic form of the governing equations while simultaneously inferring uncertainty in the system parameters. The framework combines deep symbolic regression with Gaussian process–based maximum likelihood estimation to separately model the deterministic dynamics and the noise structure, without requiring prior assumptions about their functional forms. We verify the approach on numerical benchmarks, including harmonic, Duffing, and van der Pol oscillators, and validate it on an experimental system of coupled biological oscillators exhibiting synchronization, where the algorithm successfully identifies both the symbolic and stochastic components. The framework is data-efficient, requiring as few as $10^2$-$10^3$ data points, and robust to noise---demonstrating its broad potential in domains where uncertainty is intrinsic and both the structure and variability of dynamical systems must be understood.

\end{abstract}


\section{Introduction}\label{sec:intro}

From weather prediction~\cite{lorenz2000butterfly} to modeling disease spread~\cite{tian2020investigation}, many scientific problems require uncovering relationships between variables in complex dynamical systems. Symbolic regression (SR) is a powerful technique for discovering governing equations directly from data---without assuming a specific functional form. Instead of fitting data to a predefined model, SR explores a large space of candidate mathematical expressions to find those that best describe the observed dynamics.

Approaches proposed for symbolic regression differ in the way they search the space of mathematical expressions. The sparse identification of nonlinear dynamical systems (SINDy) assumes that the governing equations can be determined from a predefined library of terms informed by \textit{a priori} knowledge of the system~\cite{Brunton2016,mangan2019model,fasel2022ensemble}. While SINDy restricts the search to sparse linear combinations of the library terms, other methodologies aim to explore the combinatorial space of expressions that arises from composing these terms with mathematical operators. For instance, genetic programming (GP) is a population-based evolutionary algorithm that builds mathematical expressions using operators inspired by evolutionary biology~\cite{gplearn,Virgolin,Virgolin_2021,burlacu2020}. In GP, equations are represented as expression trees that evolve through processes analogous to natural selection, crossover, and mutation, allowing new candidate equations to emerge. More recently, deep symbolic regression (DSR)~\cite{petersen2021deep,landajuela2022unified} uses recurrent neural networks to generate symbolic expressions directly from data. DSR treats equation discovery as a sequence-generation problem, where equations are produced token by token and guided by reinforcement learning.

Symbolic regression methodologies have been successfully applied to uncover governing equations in a wide range of systems, but they traditionally assume a deterministic world with perfect, noise-free data. Although modern methods can tolerate a certain level of noise~\cite{dascoli2023odeformer,lacava2021contemporary}, most do not explicitly quantify uncertainty.
In practice, quantifying the uncertainty associated with a discovered equation is just as important as finding the equation itself, as it directly affects safety in autonomous vehicles~\cite{tang2022prediction}, the reliability of microscopy measurements~\cite{D2NA00011C}, and the accuracy of financial forecasts~\cite{pastor2009learning}.
Moreover, in many physical systems noise is not merely an additive perturbation---it can also be \textit{multiplicative}, meaning that its amplitude depends on the state of the system~\cite{Sansa_2016}. Accurately capturing such signal-dependent noise is essential for measuring key physical quantities like mass, stiffness, and acceleration at the nanoscale~\cite{cleland,DiamagnetiqQ,MKspec}.

When randomness enters the picture, traditional ordinary differential equations (ODEs) often fall short. To model such behavior, stochastic differential equations (SDEs)~\cite{oksendal2003sde} are commonly used to describe complex phenomena influenced by intrinsic noise. SDEs extend ODEs by incorporating random fluctuations directly into the dynamics of the system through a Wiener process. In many physical systems, the noise structure is not arbitrary but reflects uncertainty in the parameters of the system---for example, fluctuations in a resonance frequency or a damping coefficient. In such cases, the diffusion term of the SDE inherits its functional structure from the drift, yielding what we refer to as SDEs with \textit{structured diffusion}. In other words, the noise in the system mirrors the structure of the deterministic dynamics, because it originates from the same physical parameters fluctuating around their nominal values. This parametric noise structure provides a natural physical motivation for constraining the diffusion term in data-driven SDE discovery.

A growing body of research has explored the use of SR in combination with nonparametric techniques to identify drift and diffusion terms in stochastic systems~\cite{tripura2023bayesian, WANG2022244, huang, Boninsegna_2018}. One widely used method is histogram-based regression (HBR)~\cite{Friedrich2011ApproachingCB, friedrich2000extracting}, which estimates local averages of the drift and diffusion coefficients by dividing the data into small bins of the state space. This approach was notably employed by Boninsegna et al.~\cite{Boninsegna_2018}, who combined HBR with SINDy to recover compact symbolic expressions for system dynamics. However, these procedures require tens of thousands~\cite{WANG2022244} to even millions~\cite{Boninsegna_2018} of data points to obtain reliable estimates. To address this, Jacobs et al.~\cite{jacobs2023hypersindy} introduced HyperSINDy, a hybrid framework that combines sparse SR with variational autoencoders to learn a compressed representation of the system, thereby enabling parameter inference without traditional binning. However, HyperSINDy is formulated as a random differential equation (RDE) rather than an SDE: noise enters through the coefficients of the ODE rather than as a separable diffusion term, so the method does not explicitly recover the functional form of the diffusion $\bs{\sigma}(\bs{X})$ (see Supplementary Material~\ref{app:rode_vs_sde}). A related approach, BISDE~\cite{tripura2023bayesian}, places Bayesian priors over the drift and diffusion coefficients to partially reduce the data requirements of HBR; however, since inference still relies on partitioning the state space into bins, it remains susceptible to the curse of dimensionality---the exponential growth in the number of required bins as the system dimension increases (see Supplementary Material~\ref{app:curse_of_dimensionality}).

Gaussian processes (GPs)~\cite{RasmussenWilliams2006} have also emerged as a compelling alternative for learning the drift and diffusion terms in SDEs~\cite{Garc_a_2017, Batz_2018}. GPs define a probability distribution over functions, enabling them to naturally capture uncertainty in both data and model predictions. By maximizing the marginal likelihood, one can fit a GP to observed data by optimizing its kernel hyperparameters. This capability makes GPs particularly effective at modeling  multiplicative noise in stochastic systems, as demonstrated by Goldberg et al.~\cite{Goldberg1998} and Kersting et al.~\cite{Kersting2007}.
However, despite their strengths in uncertainty quantification, GPs typically act as black-box models, offering little insight into the underlying structure or governing equations of a system. In contrast to symbolic regression approaches, they trade interpretability for flexibility, making them less suitable when the goal is to recover governing equations of motion.
Recent advances in neural differential equations---including Neural ODEs~\cite{chen2018neural} and Neural SDEs~\cite{li2020scalable, kidger2021neural}---have demonstrated remarkable capacity for modeling stochastic dynamics by parameterizing drift and diffusion with neural networks. Nevertheless, these approaches perform system \textit{identification} rather than equation \textit{discovery}: they learn a black-box function that reproduces the system's behavior but do not reveal the underlying equations in a form that scientists can interpret, verify, or generalize to new conditions.

In this work we introduce a modular machine learning framework for discovering stochastic nonlinear differential equations from relatively small datasets by explicitly separating the identification of deterministic and stochastic components. Our approach uses symbolic regression to discover the functional form of the drift term, while employing Gaussian processes and maximum likelihood estimation (MLE) to model complex, state-dependent noise. Unlike histogram-based regression methods, our framework does not partition the state space and is therefore free from the curse of dimensionality, achieving accurate recovery with as few as $10^2$ to $10^3$ data points.
While the framework is compatible with any symbolic regression backend, herein we demonstrate it using deep symbolic regression~\cite{petersen2021deep, landajuela2022unified}. We validate our method on a range of numerical systems---including Duffing and van der Pol oscillators---and demonstrate its effectiveness on noisy experimental data from coupled biological nonlinear oscillators. In all cases, the method successfully recovers both the governing equations and the associated noise processes in compact, interpretable symbolic form.

\section{Methodology}\label{sec:methodology}

A stochastic differential equation in the It\^o form~\cite{oksendal2003sde} takes the form
\begin{equation}
\dd{\bs{X}} = \bs{\mu} \left( \bs{X} \right) \dd{t} + \bs{\sigma} \left( \bs{X} \right) \dd{\bs{W}},
\label{eq:sde}
\end{equation}
where $\bs{X} \in \mathbb{R}^n$ is the state vector at time $t$ (e.g., displacement and velocity of a discrete system), $\bs{\mu}: \mathbb{R}^n \to \mathbb{R}^n$ is the drift term capturing the deterministic part of the system's evolution, $\bs{\sigma}: \mathbb{R}^n \to \mathbb{R}^{n \times m}$ is the diffusion term governing how random fluctuations influence the dynamics, and $\bs{W} \in \mathbb{R}^m$ is a standard Wiener process. Under the assumption that system variability arises from uncertainty in model parameters, the diffusion matrix $\bs{\sigma}(\bs{X})$ inherits its functional structure from the drift $\bs{\mu}(\bs{X})$. This structural constraint naturally captures systems where noise originates from parametric perturbations---for instance, when a stiffness coefficient or damping ratio fluctuates around its nominal value.

\begin{figure*}[t]
    \centering
    \includegraphics[width=0.88\linewidth]{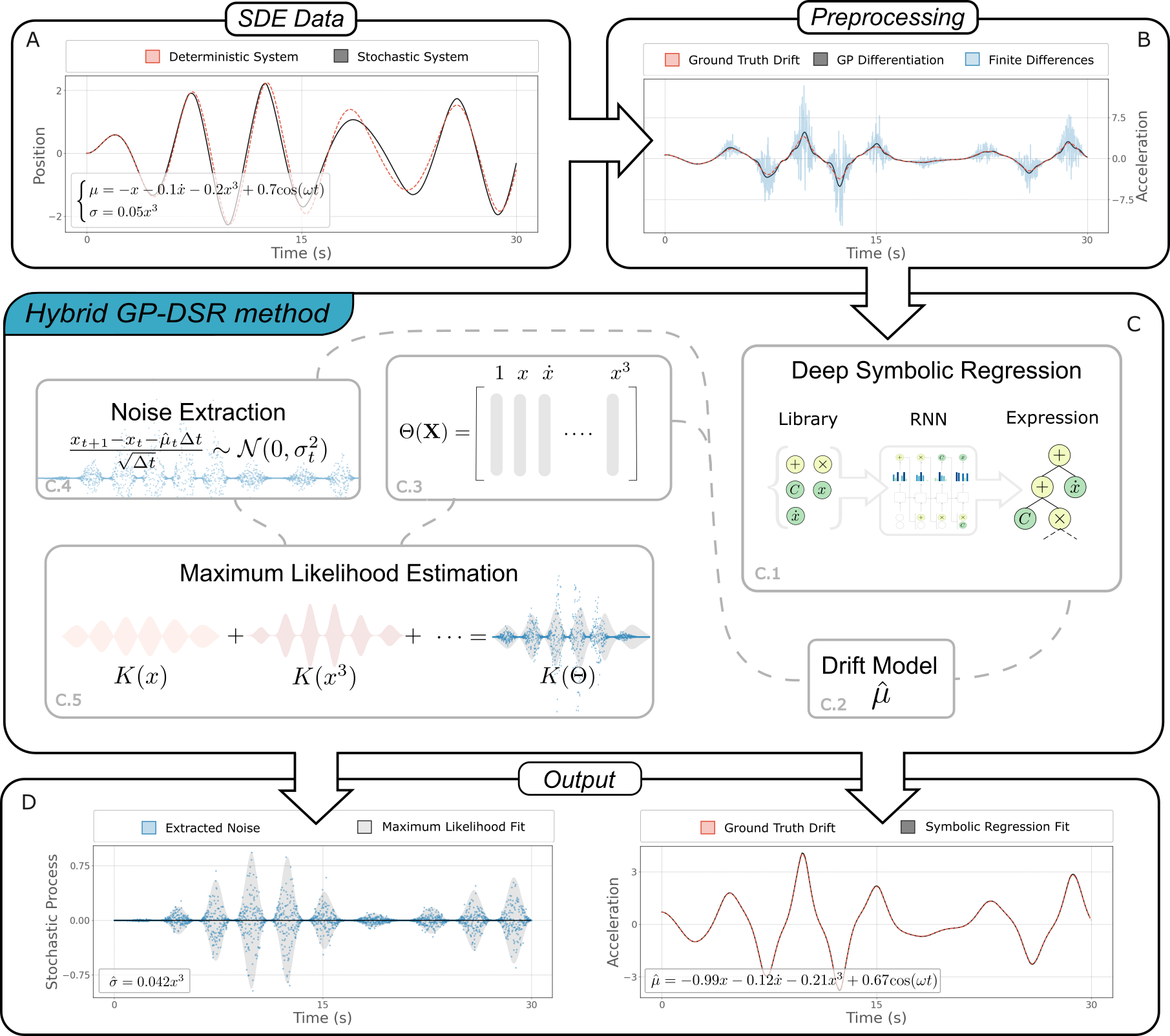}
    \caption{Overview of our methodology for discovering SDEs with structured diffusion. Here the methodology is applied to a Duffing oscillator, from data acquisition to model optimization. \textbf{A} Collect the raw data. \textbf{B} The time series data is fed into a  Gaussian process to extract the deterministic dynamics; GPs can robustly differentiate the data, as shown on the acceleration as compared to a finite finite difference scheme. \textbf{C.1-2} The preprocessed data is fed to DSR to discover the mathematical expression of the drift. \textbf{C.3-4} We build a custom dataset $\bs{\Theta}$ from the basis functions of the discovered drift and extract the noise data using the Euler--Maruyama approximation. \textbf{C.5}. We assume the noise is the sum of contributions from different components, each represented by a basis function from $\bs{\Theta}$ with its noise variance. These noise variances are optimized using MLE to best fit the stochastic process data. After optimization, variances below a certain threshold are removed to obtain a more parsimonious model without compromising predictive accuracy. \textbf{D}. The estimated drift and diffusion functions $\hat{\mu},\hat{\sigma}$ fully determined.}
    \label{fig:framework}
\end{figure*}

Our objective is to learn~\eqref{eq:sde} from a single observed time series, recovering both the symbolic form of the governing equation and the structure of uncertainty in its parameters. We first collect raw data from the system, which reflects both deterministic behavior and stochastic fluctuations. To disentangle these components, our method first denoises the time series to reveal the underlying deterministic dynamics. We then apply symbolic regression to the cleaned signal to discover the explicit functional form of the drift term $\bs{\mu}(\bs{X})$. With the deterministic model in hand, we isolate the stochastic component by subtracting the model’s prediction from the original data. Finally, we use maximum likelihood estimation with Gaussian processes to characterize the residual variability, capturing the structure and parameter uncertainty of the diffusion term $\bs{\sigma}(\bs{X})$. The different stages of the methodology are now discussed in more detail with the aid of Fig.~\ref{fig:framework}, which provides a visual overview of the framework applied to a Duffing oscillator.

The framework combines three existing building blocks in a novel way:
\begin{itemize}
    \item \textbf{Gaussian processes}~\cite{RasmussenWilliams2006} for denoising the time series and estimating state derivatives;
    \item \textbf{Symbolic regression}~\cite{petersen2021deep} for discovering the functional form of the drift term $\bs{\mu}(\bs{X})$;
    \item \textbf{Maximum likelihood estimation with structured GP kernels} for inferring the diffusion term $\bs{\sigma}(\bs{X})$ from the residuals.
\end{itemize}
The novel contributions of this work are: (i) the \textit{structured diffusion} assumption---that the noise inherits its functional form from the drift---which physically constrains the diffusion and enables accurate inference from small datasets; (ii) the use of MLE with state-dependent GP kernels specifically designed to recover the symbolic structure of $\bs{\sigma}(\bs{X})$; and (iii) the modular architecture of the pipeline, which allows the symbolic regression component to be replaced by any alternative SR method.

\paragraph{Preprocessing} In this phase (see \hyperref[fig:framework]{Fig.~\ref*{fig:framework}B}) we use the time series data to estimate the drift function $\bs{\mu}(\bs{X})$ and the diffusion term $\bs{\sigma}(\bs{X})$ in~\eqref{eq:sde}. This requires accurately differentiating the state trajectories, which is challenging due to fluctuations induced by the noise. To address this we employ Gaussian processes, which yield smooth and robust estimates of the derivatives. This is shown in the acceleration curves of \hyperref[fig:framework]{Fig.~\ref*{fig:framework}B}, where GP differentiation is compared to a finite difference scheme, the latter showing highly oscillatory and unreliable results.

\paragraph{Symbolic regression of the drift term}
To obtain a parsimonious expression from the preprocessed time series data, we employ a symbolic regression algorithm (see \hyperref[fig:framework]{Fig.~\ref*{fig:framework}C.1}). In particular, we use deep symbolic regression~\cite{petersen2021deep}, which has shown strong empirical performance in recent benchmarks~\cite{lacava2021contemporary}.
DSR uses a recurrent neural network (RNN) trained via reinforcement learning to generate candidate mathematical expressions as sequences of tokens---such as operators, functions, and constants---drawn from a user-specified library. DSR can flexibly combine these tokens into a wide variety of symbolic expressions, exploring a richer space of candidate equations. For full algorithmic details, see Supplementary Material~\ref{app:dsr}.

\paragraph{Diffusion term}
After discovering the symbolic expression representing the deterministic dynamics of the system, we proceed to infer the structure of the diffusion term. The key insight is straightforward: if noise originates from uncertain parameters, then the noise structure mirrors the structure of the drift equation. For instance, if the stiffness coefficient fluctuates, the resulting noise will be proportional to the same terms that the stiffness multiplies in the drift. We formalize this by modeling the noise as a combination of independent normal distributions, each associated with a distinct basis function of the drift equation (see Supplementary Material \ref{app:GP}). To represent these basis functions, we construct nonlinear combinations of the states $\bs{X}$, forming a tailored feature set denoted as $\bs{\Theta}$ (see \hyperref[fig:framework]{Fig.~\ref*{fig:framework}C.3}). For instance, for the Duffing oscillator shown in Fig.~\ref{fig:framework}, the discovered drift includes terms proportional to $x$, $\dot{x}$, and $x^3$, so the feature set $\bs{\Theta}$ contains the corresponding columns $[1, x, \dot{x}, x^3, \ldots]$, where each column represents a candidate basis function for the diffusion (see Supplementary Material for full details).

Each basis function is associated with noise through the covariance kernel
\begin{equation}
    \bs{K}_i(\bs{\Theta}_i, \bs{\Theta}_i') = (\bs{\Theta}_i \cdot \bs{\Theta}_i') \cdot \sigma_i^2 \, \delta(\bs{\Theta}_i - \bs{\Theta}_i'),
\end{equation}
where $\sigma_i^2$ is the noise variance associated with the $i$th basis function, and the Kronecker delta $\delta$ enforces statistical independence between noise realizations at different time points---a standard white-noise assumption. In practice, this kernel reduces to a diagonal covariance matrix whose entries scale with the squared values of the basis functions, so that the noise amplitude at each time point is modulated by the local value of the corresponding state-dependent term. The overall noise covariance matrix $\bs{K}(\bs{\Theta})$ is then obtained by summing these individual contributions, i.e.,
\begin{equation}
    \bs{K}(\bs{\Theta}) = \sum_{i=1}^N \bs{K}_i(\bs{\Theta}_i),
\end{equation}
where \(N\) is the number of basis functions present in the drift equation. This formulation enables us to quantify the contribution of each basis function to the total noise in the system. In addition to capturing parameter uncertainty, we also consider the possibility of additive noise acting directly on the system dynamics.

To extract the stochastic component from the data, we apply the Euler--Maruyama scheme~\cite{kloeden1999numerical} (see \hyperref[fig:framework]{Fig.~\ref*{fig:framework}C.4}). Once the deterministic dynamics are known and subtracted from the observed signal, the remaining residual consists purely of noise whose variance reveals the diffusion amplitude. The discretization reads
\begin{equation}
    \bs{X}_{t+1} = \bs{X}_t + \bs{\mu}(\bs{X}_t) \, \Delta t + \bs{\sigma}(\bs{X}_t) \, \Delta\bs{W}_t
    \label{eq:euler}
\end{equation}
where $\Delta\bs{W}_t = \bs{W}_{t+\Delta t} - \bs{W}_t \sim \mathcal{N}(\bs{0}, \Delta t \, \bs{I}_n)$ is the Wiener increment over the time step $\Delta t$, consistent with the $n$-dimensional formulation of~\eqref{eq:sde}. Since $\Delta W_t \sim \mathcal{N}(0, \Delta t)$ for each scalar component, dividing the stochastic increment by $\sqrt{\Delta t}$ yields a standard normal. We therefore define a residual for the $i$th component of~\eqref{eq:euler} that isolates the stochastic part as
\begin{equation}
    \frac{X_{t+1} - X_{t} - \hat{\mu}_{t} \Delta t}{\sqrt{\Delta t}} \sim \mathcal{N} \left( 0,  \sigma_{t}^2 \right),
    \label{eq:extraction}
\end{equation}
where we left out the index $i$ for conciseness. Here $\hat{\mu}_t$ denotes the drift \textit{estimated} by symbolic regression, as opposed to the true drift $\bs{\mu}(\bs{X}_t)$ in~\eqref{eq:euler}: the hat distinguishes the model discovered from data from the unknown ground-truth function.

After identifying the drift term $\hat{\bs{\mu}}(\bs{X})$ through symbolic regression, we substitute the discovered model into the discretized SDE to extract the residual noise from the data. These residuals correspond to the stochastic component $\bs{\sigma}(\bs{X})\,\Delta\bs{W}$, obtained by subtracting the inferred drift from the observed state increments. Importantly, while the preprocessing stage uses Gaussian process derivatives to obtain a smooth signal for equation discovery, the noise extraction step relies on finite differences to preserve the stochastic content of the data (see Supplementary Material~\ref{app:gpdenoising} for details).

To estimate the variance structure of the noise, we optimize the parameters \(\{\sigma_i^2\}\) defining the individual covariance kernels $\bs{K}_i(\bs{\Theta}_i)$ using MLE. Specifically, we minimize the negative log-likelihood of a multivariate Gaussian model with zero mean and covariance $\bs{K} (\bs{\Theta} ) = \sum_i \bs{K}_i(\bs{\Theta}_i)$:
\begin{equation}
-\log \mathcal{L} = \frac{1}{2} \bs{y}^\intercal \bs{K} (\bs{\Theta})^{-1} \bs{y} + \frac{1}{2} \log \left| \bs{K} (\bs{\Theta}) \right| + \frac{p}{2} \log (2\pi),
\end{equation}
where $\bs{y}$ is the vector of extracted residuals and $p$ is the number of data points. The first term quantifies the data fit, measuring the discrepancy between the observed residuals and the predicted distribution from the model. After the MLE step, the coefficients of the diffusion term are derived from the optimized variance parameters associated with each $\bs{K}_i(\bs{\Theta}_i)$ (see \hyperref[fig:framework]{Fig.~\ref*{fig:framework}D}).
Finally, to promote parsimony, we prune noise components whose estimated variance falls below a threshold---thereby effectively removing basis functions that contribute negligibly to the diffusion. This approach, known as automatic relevance determination (ARD), simplifies the model while preserving its predictive power (see \hyperref[fig:framework]{Fig.~\ref*{fig:framework}D}).

\section{Verification on Numerical Datasets}
\label{sec:res}

We first verified our framework on numerical datasets generated from oscillators, including a linear harmonic oscillator, a van der Pol oscillator, and a Duffing oscillator, each subjected to Gaussian noise in different parameters. All systems are discretized using the Euler--Maruyama scheme over \SI{50}{\sec}, resulting in a dataset of \num{5000} points per trajectory (see Supplementary Material~\ref{app:euler}). The verification strategy progresses from additive noise in the linear oscillator, to multiplicative noise, then to nonlinear dynamics with structured diffusion, and finally to scenarios where noise affects all parameters.

\paragraph{Linear oscillator with additive noise.}
The first numerical experiment considers the linear harmonic oscillator with additive noise:
\begin{equation}
\left\{
\begin{aligned}
    &\mu = -\omega_0^2 x - 2 \zeta \omega_0\dot{x} + f \cos{\Omega t}, \\
    &\sigma = \sigma_a, \\
    &\omega_0 = \Omega = \SI[per-mode=symbol]{1}{\radian\per\second}, \quad \zeta = 0.1 ,\quad \text{and} \quad f = \SI[per-mode=symbol]{0.4}{\newton\per\kilogram},
\end{aligned}
\right.
\label{eq:lin_a}
\end{equation}
where $\omega_0$ denotes the resonance frequency, $\zeta$ the damping ratio, $f$ the amplitude of the external driving force exciting the system at resonance, and $\sigma_a$ the intensity of the additive noise.

\begin{figure*}[t]
    \centering
    \includegraphics[trim={0 0 0 0},clip,width=0.9\linewidth]{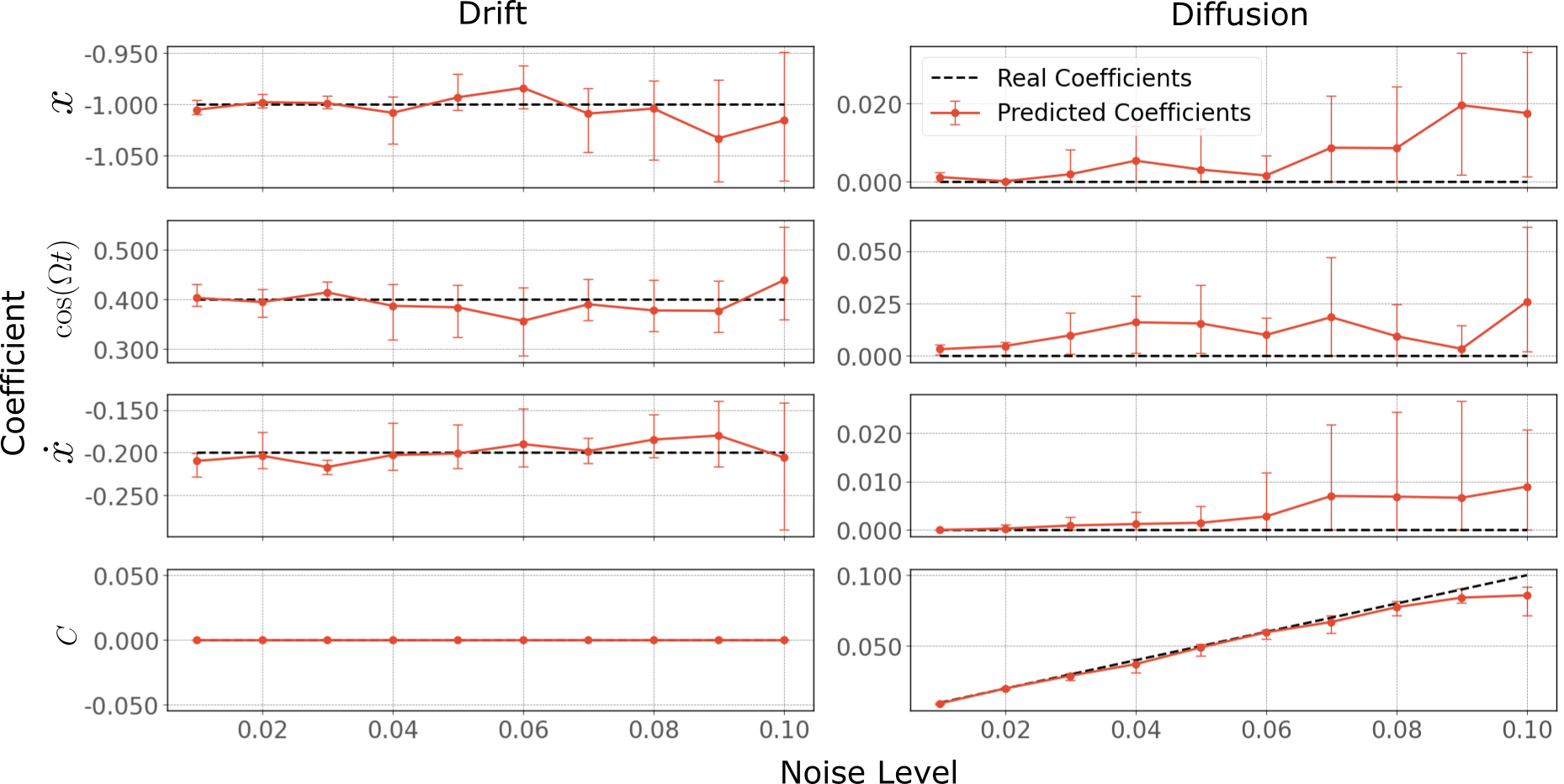}
    \caption{\textbf{Linear oscillator with additive noise.}
Predicted coefficients for the drift (left column) and diffusion (right column) as a function of noise level. Red lines indicate the model predictions, while black dashed lines denote the ground truth.
Error bars (where shown) correspond to the 2.5\%--97.5\% quantiles across realizations, reflecting the framework's uncertainty when applied to five independent datasets obtained from integrations of the same system.
The framework consistently preserves structural accuracy across noise levels and yields precise estimates of the diffusion amplitude, although spurious diffusion activations appear at higher noise intensities.
In the diffusion panel, $C$ denotes the additive noise constant $\sigma_a$ (see~\eqref{eq:lin_a}).
}
    \label{fig:lin_a}
\end{figure*}

We systematically vary the noise magnitude from $0\%$ to $10\%$ (defined relative to the nominal parameter values; see Supplementary Material~\ref{app:resultstats} for details) to evaluate its effect on the prediction accuracy of drift and diffusion coefficients. The results are summarized in Fig.~\ref{fig:lin_a}, which shows the predicted coefficients of drift and diffusion for increased noise levels. As apparent from the figure, the framework preserves the correct functional form across the entire noise sweep. However, as noise level is increased, small artificial noise contributions appear in terms that should be zero (see predicted coefficients for the diffusion in Fig.~\ref{fig:lin_a}).

\begin{figure}[t]
    \centering
    \includegraphics[width=0.6\linewidth]{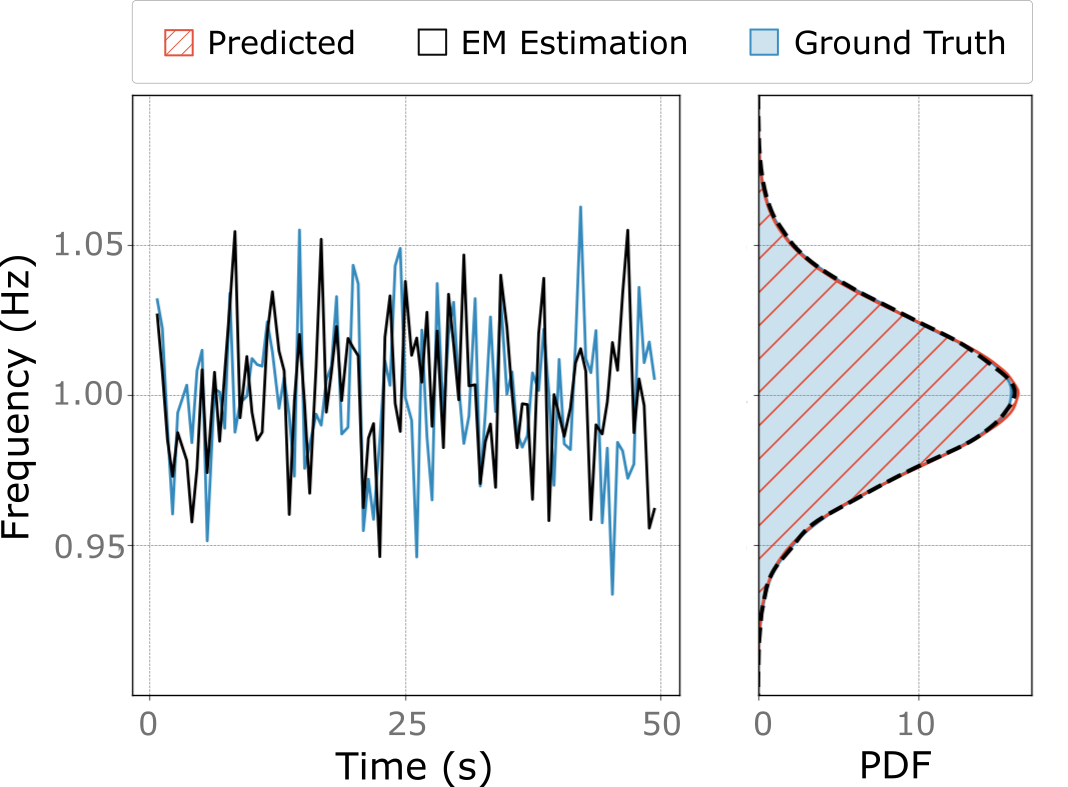}
    \caption{\textbf{Multiplicative frequency noise.}
Estimation of the natural frequency distribution for the linear system with $5\%$ frequency noise.
\emph{Left:} Ground truth frequency variations (blue) compared with estimated variations (black) using the Euler--Maruyama approximation with $\Delta t = 0.01$.
\emph{Right:} Ground truth probability density function of the frequency parameter (blue), approximated distribution from the stochastic process data (black dashed), and MLE-predicted distribution (red shaded).}
    \label{fig:lin_w}
\end{figure}

Our second numerical experiment introduces multiplicative frequency noise in the same linear oscillator, modeled as $\sigma_w = 2\omega_0\Delta\omega_0$, with the diffusion process being $\sigma_w x$. Figure~\ref{fig:lin_w} illustrates the recovery of the resonance frequency distribution under a $5\%$ frequency noise level. While the Euler--Maruyama approximation does not fully resolve the time-domain variations (Fig.~\ref{fig:lin_w}, left), it nevertheless captures the overall distribution of the frequency noise (Fig.~\ref{fig:lin_w}, right). The recovery is further consolidated by the MLE step.

\begin{figure*}[t]
    \centering
    \includegraphics[width=0.9\textwidth]{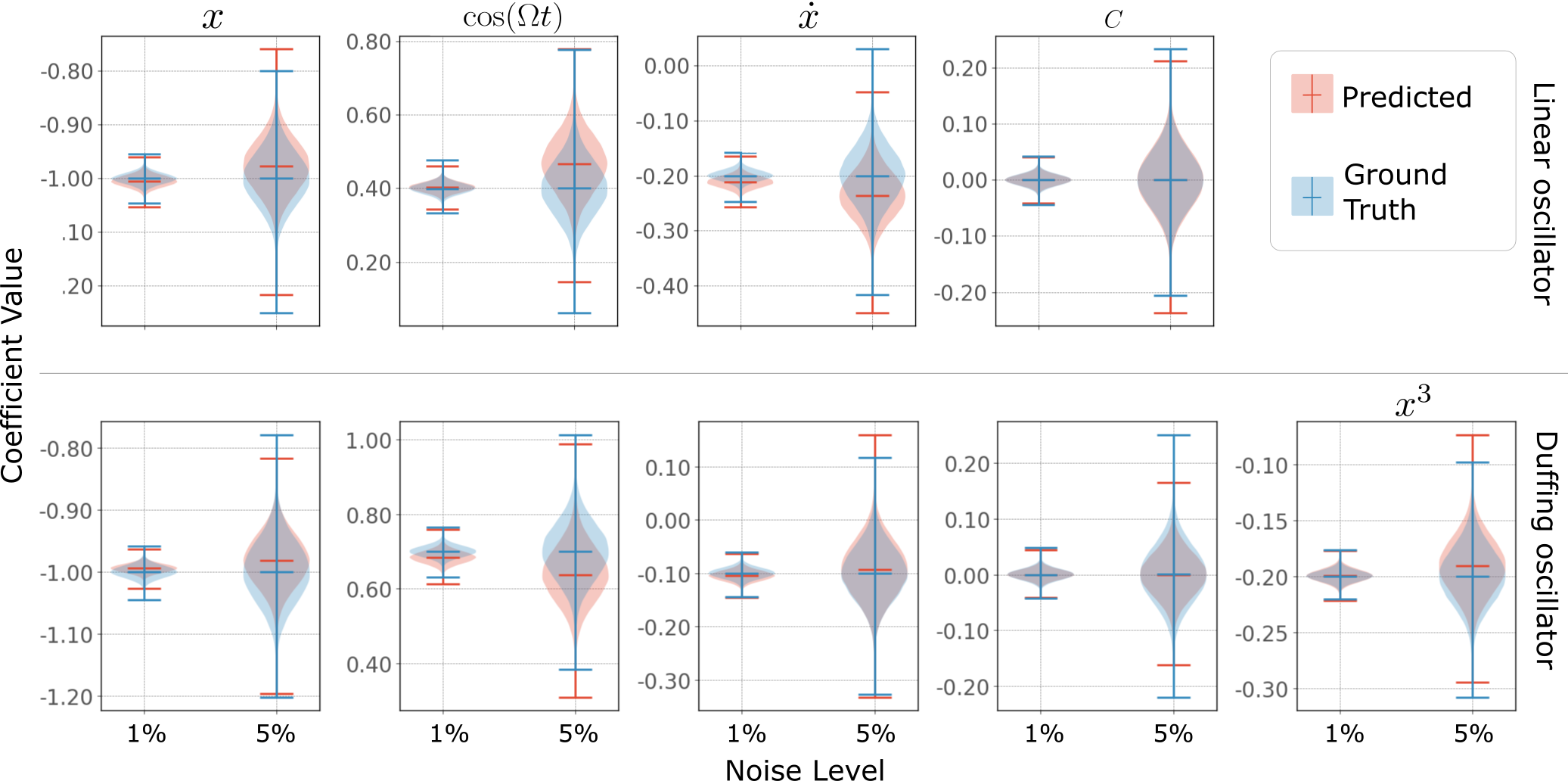}
    \caption{\textbf{Linear and Duffing oscillators with noise in all parameters.}
Violin plots show distributions of recovered coefficients at $1\%$ and $5\%$ noise.
Blue bars denote ground truth, while red bars indicate predictions.
Each distribution is modeled as Gaussian, with the mean corresponding to the drift and the standard deviation to the diffusion of the coefficients.
In both panels, $C$ denotes the additive noise constant $\sigma_a$ (see~\eqref{eq:lin_all}).}
    \label{fig:res}
\end{figure*}

To further increase the complexity, we simulate both a linear harmonic and a Duffing oscillator with noise affecting every parameter. For the linear harmonic oscillator we use the same parameters as in \eqref{eq:lin_a}, with a more complex diffusion process:
\begin{equation}
    \sigma = \sigma_a + \sigma_w x + \sigma_d \dot{x} + \sigma_f u.
    \label{eq:lin_all}
\end{equation}
where $\sigma_a$ is the additive noise intensity, $\sigma_w$, $\sigma_d$, and $\sigma_f$ are the noise amplitudes associated with the state $x$, the velocity $\dot{x}$, and the external forcing $u$, respectively.
Our Duffing oscillator is described by
\begin{equation}
\left\{
    \begin{aligned}
        &\mu = -\omega_0^2 x - d \dot{x} - k_3 x^3 + fu, \\
        &\omega_0 = 1\frac{\si{\radian}}{\si{\second}}, \, d = \frac{0.1}{\si{\second}},\, k_3= \frac{0.2}{\si{\metre\squared\second\squared}}, \text{~and~} f = \SI[per-mode=symbol]{0.7}{\newton\per\kilogram},
    \end{aligned}
\right.
\end{equation}
where compared to \eqref{eq:lin_all}, the diffusion has an additional multiplicative noise term in the nonlinear stiffness. For numerical both experiments, noise levels are set to $1\%$ and $5\%$. As shown in Fig.~\ref{fig:res}, the framework accurately captures both the drift and diffusion coefficients, with estimated noise amplitudes matching the ground truth values---confirming the robustness of the framework even for highly noisy systems.

We also applied the framework to a van der Pol oscillator with structured diffusion, where the method successfully recovered both the drift and diffusion terms (see Supplementary Material~\ref{app:resultstats}).

\section{Experimental Data on Coupled Biological Oscillators}
\label{sec:coupled_oscillators}

Having established the framework's accuracy on controlled numerical benchmarks, we now demonstrate its effectiveness on experimental data with limited samples. Specifically, we examine the synchronization phenomenon between two biological oscillators. Synchronization---the process by which interacting oscillators adjust their rhythms to operate in unison---is observed across nature, from flocking birds~\cite{birdflock2014} to flashing fireflies~\cite{buckfireflies1976}.

\begin{figure}[t]
    \centering
    \includegraphics[width=0.5\linewidth]{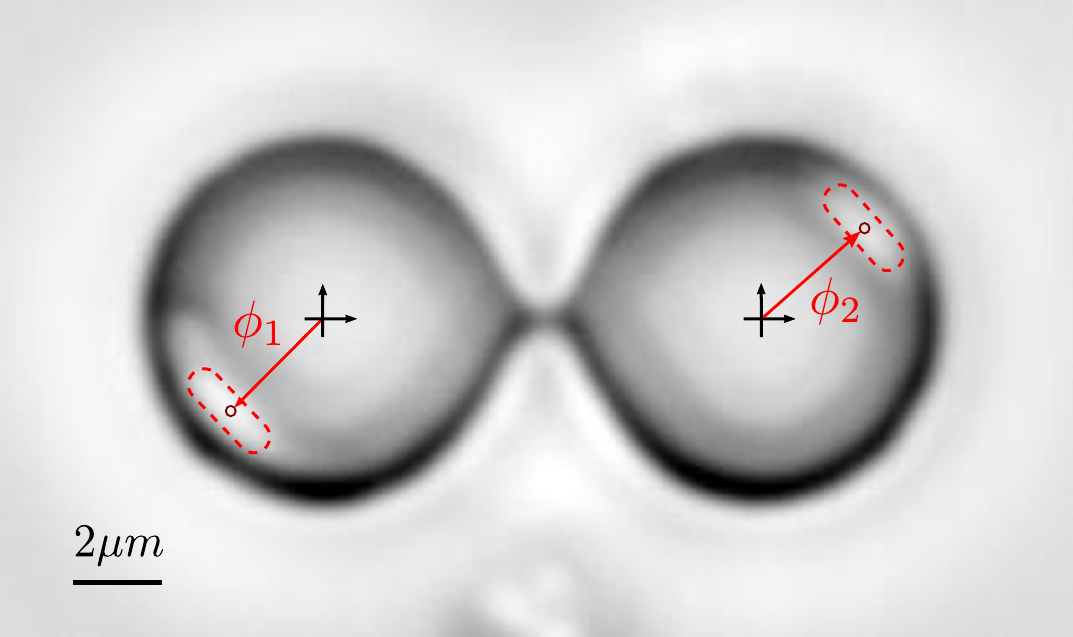}
    \caption{\textbf{The experiment setup.} Microscope image of two \textit{E. coli} bacteria trapped in separate cavities connected by a narrow channel~\cite{japaridze2024synchronization}.}
    \label{fig:bacteria_schematic}
\end{figure}

We apply our method to experimental data from Japaridze et al.~\cite{japaridze2024synchronization}, where synchronization was observed between two \textit{Escherichia coli} cells in separate microcavities connected by a narrow channel (Fig.~\ref{fig:bacteria_schematic}). The datasets were obtained by tracking the angular position of each bacterium from low-frame-rate microscope video, resulting in two sparsely and irregularly sampled time series of approximately $10^2$ data points each.

As derived in Supplementary Material~\ref{app:adler}, the phase dynamics of coupled bacterial oscillators can be reduced to the noisy Adler equation:
\begin{equation} \label{eq:adler}
    \frac{d\varphi}{dt} = \Delta\omega - K \sin(\varphi) + \xi(t)
\end{equation}
where $\varphi = \phi_2 - \phi_1$ is the phase difference, $\Delta\omega$ the frequency mismatch, $K$ the coupling strength, and $\xi(t)$ the additive noise from bacterial velocity fluctuations. In Japaridze et al.~\cite{japaridze2024synchronization}, the parameters $\Delta\omega$ and $K$ were estimated from the statistical properties of this equation.

\begin{figure*}[ht!]
    \centering
    \includegraphics[width=0.95\textwidth]{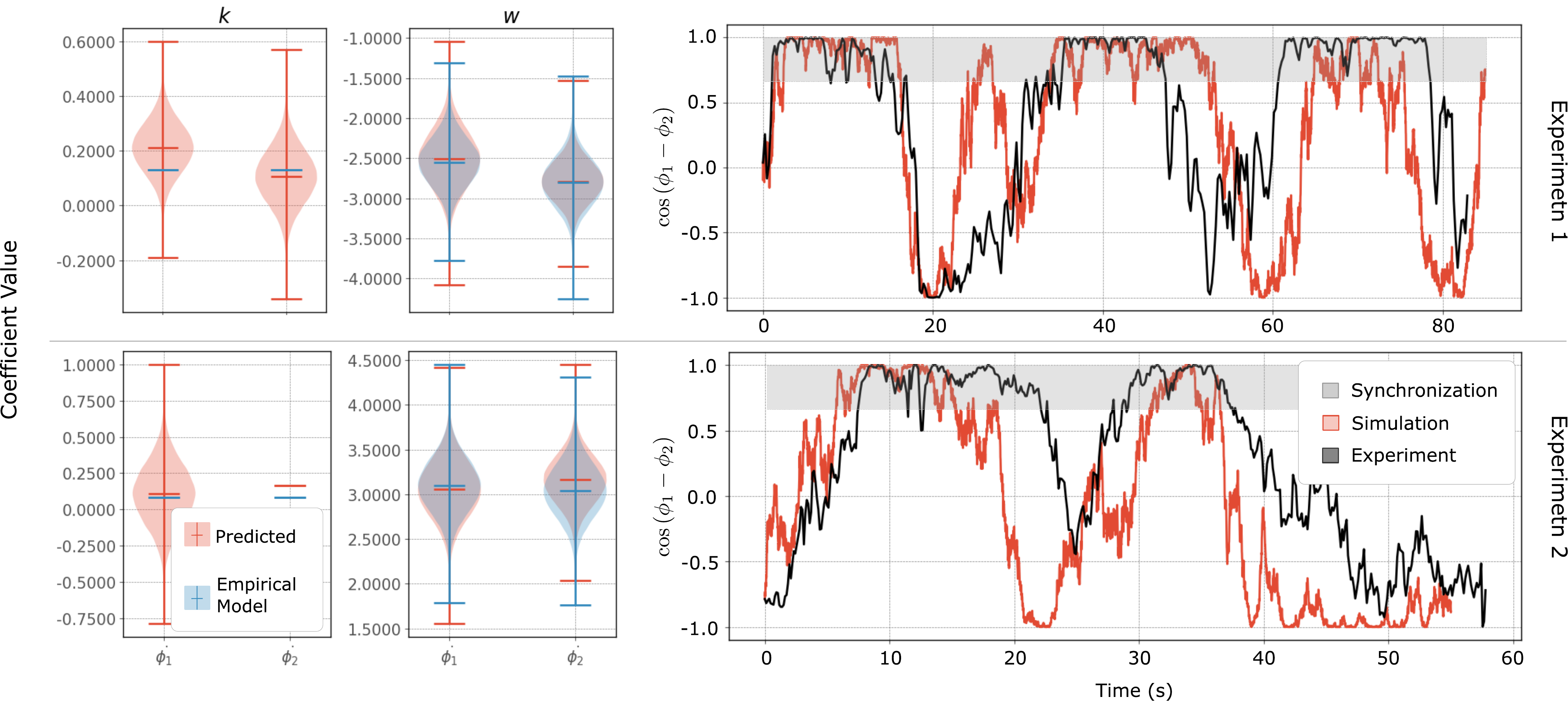}
    \caption{\textbf{Comparison between experimental data and discovered models for two synchronization datasets.}
\emph{Left:} Violin plots show the distributions of coupling strength ($k$) and angular velocity ($\omega$) for the two oscillators. Blue bars represent empirical estimates obtained with the method of Japaridze et al.~\cite{japaridze2024synchronization}, while red bars indicate predictions from our algorithm. Each distribution is modeled as Gaussian, with the mean corresponding to the drift and the standard deviation to the diffusion of the coefficients.
\emph{Right:} Cosine of the phase difference $(\phi_1 - \phi_2)$ over time, comparing simulations of the discovered model with experimental data. The gray-shaded region marks the interval during which synchronization occurs.}
    \label{fig:bacteria_osc}
\end{figure*}

\paragraph{Application of the framework.}
Rather than regressing the phase difference directly, we applied our algorithm to the individual phases of each bacterium, allowing the framework to extract richer information from the data. The recovered models closely resemble~\eqref{eq:adler}, and the estimated drift and diffusion coefficients align with those reported by Japaridze et al.~\cite{japaridze2024synchronization} (Fig.~\ref{fig:bacteria_osc}, left).

Notably, the framework reveals two findings beyond the original analysis. First, in addition to additive noise, it predicts noise in the coupling parameter $K$ (see Supplementary Material~\ref{app:em}). Second, it identifies asymmetric coupling strengths between the two oscillators---a physically plausible result, since the bacteria have different natural frequencies and the thrust force of an \textit{E.~coli} cell is proportional to its velocity~\cite{Elgeti_2015}. This contrasts with the symmetric coupling assumed in~\cite{japaridze2024synchronization}. The discovered models, when simulated, accurately reproduce the experimental dynamics, including the frequency and duration of synchronization events (Fig.~\ref{fig:bacteria_osc}, right).

To benchmark our framework against the state of the art, we compared our results with BISDEs~\cite{WANG2022244}, a Bayesian method for SDE discovery. Despite being provided with a library of trigonometric functions designed to make the problem comparably challenging (see Supplementary Material~\ref{app:bact} for full details), BISDEs failed to recover the Adler equation from either dataset. We attribute this primarily to the limited amount of available data: BISDEs typically requires substantially larger datasets for reliable inference, whereas our framework operates effectively with the sparse, irregularly sampled data characteristic of biological experiments.

\section{Discussion}\label{sec:discussion}

The proposed framework demonstrates that combining symbolic regression with Gaussian process-based maximum likelihood estimation enables the recovery of both the governing equations and the noise structure of stochastic dynamical systems from relatively small datasets. The verification on numerical benchmarks and the experimental validation on the synchronization of biological oscillators confirm the accuracy of the method and data efficiency. We now discuss the main observations, limitations, and potential directions for future research.

The maximum likelihood estimation step reliably identified the basis functions that contribute most significantly to the stochastic process. However, some spurious activations were also observed: when the MLE jointly optimizes all noise variances $\{\sigma_i^2\}_{i=1}^N$, basis functions that do not truly contribute to the diffusion may still be assigned small non-zero variances; this is because finite-sample correlations between the residuals and irrelevant terms prevent the optimizer from converging exactly to zero for those components. In our experiments, while we intentionally held the ARD threshold constant across all cases to clearly expose these effects, in practice the threshold should be adapted to the noise level of the system under study.

Our focus has been on demonstrating the effectiveness of SR-based methods in real-world noisy systems---a setting that remains largely unexplored---rather than on benchmarking symbolic regression recovery rates, which have been studied extensively in prior work~\cite{lacava2021contemporary,dascoli2023odeformer,petersen2021deep}.
While DSR was used here as the SR engine, the framework is modular and can be replaced with other SR methods, allowing advances in equation discovery to translate directly into better SDE inference.
It is worth noting, however, that the effectiveness of the framework depends on domain-informed choices, such as the library of tokens and constraints on the search space; when symbolic regression fails to recover the correct symbolic form, the Gaussian process fit may still capture the data accurately---effectively reducing the method to system identification rather than equation discovery. These cases are documented in Supplementary Material~\ref{app:additionalresults}.

Our approach assumes that the diffusion term can be expressed using the same basis functions as the drift---a modeling choice we refer to as \textit{structured diffusion}. This assumption is well suited to systems in which noise arises from parametric uncertainty, such as fluctuations in stiffness, damping, or coupling constants. However, systems whose noise structures unrelated to the deterministic dynamics would require extending the framework beyond its current scope. This represents a trade-off between flexibility and data efficiency: the structured assumption enables accurate inference from very limited data, whereas methods that allow arbitrary diffusion forms typically require substantially larger datasets~\cite{WANG2022244, Nabeel_2023, Boninsegna_2018}.

A practical consideration is the sensitivity of the Euler--Maruyama residual extraction to the discretization time step $\Delta t$; as $\Delta t$ increases relative to the system's characteristic timescale, the estimated diffusion amplitudes degrade. This effect is analyzed in detail in Supplementary Material~\ref{app:em}, where we show that the approximation remains accurate for the time steps used in our experiments, and where we identify the regime in which it breaks down. Guidance on setting the remaining hyperparameters---including the ARD threshold and the DSR token library---is provided in Supplementary Material~\ref{app:numerical_settings} and~\ref{app:additionalresults}.

Looking ahead, several extensions are natural. Combining our framework with non-parametric diffusion estimation could enable the discovery of SDEs with arbitrary noise structures. Incorporating temporal GP kernels could capture colored noise processes, moving beyond the white-noise assumption. The successful application to coupled biological oscillators with sparse, irregularly sampled data suggests broader applicability to experimental systems in biophysics and nanomechanics---domains where noise quantification is critical and data collection is costly, and where the data efficiency of the framework offers a practical advantage over histogram-based methods.

Code and data are available at \url{https://github.com/matteobosso/AGP-SDE}.

\section{Acknowledgments}
This work was partially supported by the European Union through the ERC Consolidator Grant NCANTO (No. 101125458). A.~M.~Aragón and M.~Theulings further acknowledge support from the NGF-AiNed XS Europe 2023 grant (No. NGF.1609.23.007), awarded by the Nederlandse Organisatie voor Wetenschappelijk Onderzoek (NWO).

\appendix

\section{Maximum likelihood estimation for the inference of the stochastic process}
\label{app:GP}
\begin{figure}[t]
    \centering
    \includegraphics[width=0.75\linewidth]{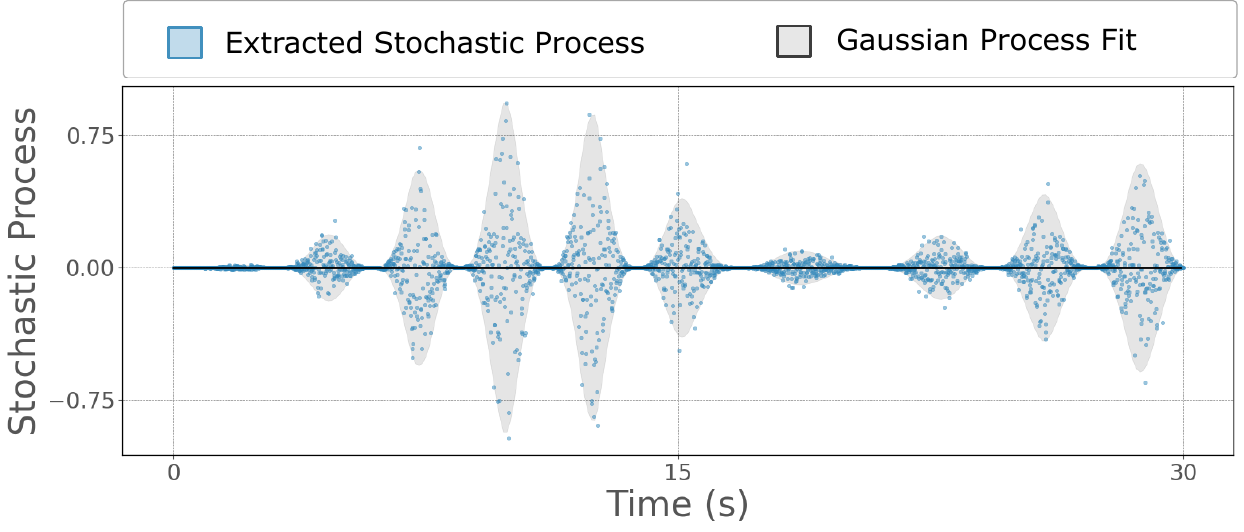}
    \caption{The extracted noisy process (blue dots) is plotted against the fit of our tailored GP (shaded grey area). Our method is capable of fitting this heteroscedastic noise with remarkable precision.}
    \label{fig:noise}
\end{figure}

When extracting stochastic process data using the Euler--Maruyama residual (Eq.~\eqref{eq:extraction} of the main text), one may encounter complex types of noise (see Figure \ref{fig:noise}). Under the assumption that the diffusion shares the same basis functions as the drift (i.e., noise enters through parametric uncertainty), each basis function contributes its own noise to the system, and the overall noise is the sum of these contributions.

\noindent To accurately represent this noise, we optimize the variances of each basis function using maximum likelihood estimation (MLE) to best fit the extracted noise data. Gaussian processes (GPs) provide a suitable framework for this optimization. GPs can model additive noise using the white noise kernel in their standard formulation:
\[
k_{\text{WN}}(\bs{X}, \bs{X}') = \sigma_n^2 \delta(\bs{X} - \bs{X}')
\]
where \(\sigma_n\) is the standard deviation of the noise observed in the data.

\noindent Thus, the variability in certain model parameters results in noise that is multiplicative with the basis functions present in the drift. These can be represented by a nonlinear function matrix similarly to the library of basis functions in the sparse identification of nonlinear dynamics (SINDy)~\cite{Brunton2016}:
\begin{equation*}
    \bs{\Theta}(\bs{X}) :=
    \begin{bmatrix}
        \bs{1} & \bs{X} & \bs{X}^2 & \bs{X}^3 & \cdots
    \end{bmatrix},
\end{equation*}
where $\bs{1} \in \mathbb{R}^m$ denotes a column vector of ones, and $\bs{X}^k$ denotes the matrix of all $k$th-order polynomial interaction terms in the state variables.
The matrix $\bs{X}^k$ collects all $k$th-order polynomial interaction terms. For example, $\bs{X}^2$ contains all quadratic interactions:
\begin{equation*}
\bs{X}^{2} :=
\begin{bmatrix}
x_1^2(t_1) & x_1(t_1)x_2(t_1) & \cdots & x_i(t_1)x_j(t_1) & \cdots & x_n^2(t_1) \\
x_1^2(t_2) & x_1(t_2)x_2(t_2) & \cdots & x_i(t_2)x_j(t_2) & \cdots & x_n^2(t_2) \\
\vdots & \vdots & \ddots & \vdots & \ddots & \vdots \\
x_1^2(t_m) & x_1(t_m)x_2(t_m) & \cdots & x_i(t_m)x_j(t_m) & \cdots & x_n^2(t_m)
\end{bmatrix}.
\end{equation*}
Leveraging the drift expression identified by symbolic regression, we retain only the basis functions present in that expression, thereby significantly reducing the number of columns in $\bs{\Theta}$.
Rather than composing nonlinear kernels over the raw state variables $\bs{X}$, we supply the evaluated basis functions as GP inputs directly. This reduces the kernel structure to simple linear (dot-product) interactions between the columns of $\bs{\Theta}$.
Since the nonlinear structure is already encoded in the columns of $\bs{\Theta}$---which may include functions such as $\sin, \cos, \exp, \log$---the GP kernel only needs to model linear covariance and additive noise contributions.
Each kernel for each column of $\bs{\Theta}$ is
\begin{equation}
  \bs{K}_i(\bs{\Theta}_i, \bs{\Theta}_i') = (\bs{\Theta}_i \cdot \bs{\Theta}_i') \times \sigma_{i}^2 \delta(\bs{\Theta}_i - \bs{\Theta}_i'), \notag
\end{equation}
where $\sigma_{i}^2$ denotes the noise variance of the $i$th basis function in $\bs{\Theta}$ and $\delta$ the Kronecker delta function.
The aggregate kernel for the dataset \(\bs{\Theta}\) is then obtained by summing these individual kernels
\begin{equation}
    \bs{K}(\bs{\Theta}) = \sum_{i=1}^N \bs{K}_i(\bs{\Theta}_i,\bs{\Theta}_i'),
    \notag
\end{equation}
where \(N\) is the number of columns in \(\bs{\Theta}\) (the number of basis functions present in the drift). Additionally, we investigate the presence of additive noise in the system by adding a white noise kernel to the aggregate kernel above.

The GP is then trained with $\bs{\Theta}$ as input and the extracted stochastic process data (Fig.~\ref{fig:noise}) as output.
Denoting by $\bs{y} \in \mathbb{R}^m$ the vector of extracted noise values ($m$ is the number of observations), the algorithm optimizes the noise parameters of the white kernels to minimize the negative log-likelihood
\begin{equation*}
-\log(\mathcal{L}) = \frac{1}{2} \bs{y}^\intercal \bs{K}(\bs{\Theta}, \bs{\Theta})^{-1} \bs{y} + \frac{1}{2} \log  \left( \left| \bs{K}(\bs{\Theta}, \bs{\Theta}) \right| \right) + \frac{m}{2} \log (2 \pi).
\end{equation*}
The likelihood noise $\sigma_n^2 \bs{I}$ (where $\bs{I}$ is the $m \times m$ identity matrix) within the Gaussian process is determined by the summation of contributions from different kernels. The modulation provided by the Kronecker delta function ensures that noise contributions are only considered when $x_i = x_i'$ (using the standard GP convention that $x_i'$ denotes a second evaluation point), affecting only the diagonal elements of the covariance matrix. The mathematical expression for the likelihood noise with our tailored kernel becomes
\begin{equation*}
    \sigma_n(\bs{\Theta})^2 = \sum_{i=1}^{N} \bs{\Theta}_i^2 \cdot \sigma_{i,n}^2.
\end{equation*}
As a result, the noise covariance matrix will be diagonal, with each diagonal entry varying according to the specific squared value of each \(\bs{\Theta}_i\) from \(\bs{\Theta}(\bs{X})\).

\section{Deep symbolic regression}
\label{app:dsr}

Deep symbolic regression (DSR)~\cite{petersen2021deep} is a machine learning approach that is used to discover closed-form mathematical expressions from data by framing the search over symbolic expressions as a sequential decision-making problem. We provide here a self-contained description of the algorithm for completeness. For further details, we refer the reader to~\cite{petersen2021deep, landajuela2022unified}.

\begin{figure}[t]
    \centering
    \includegraphics[width=0.85\linewidth]{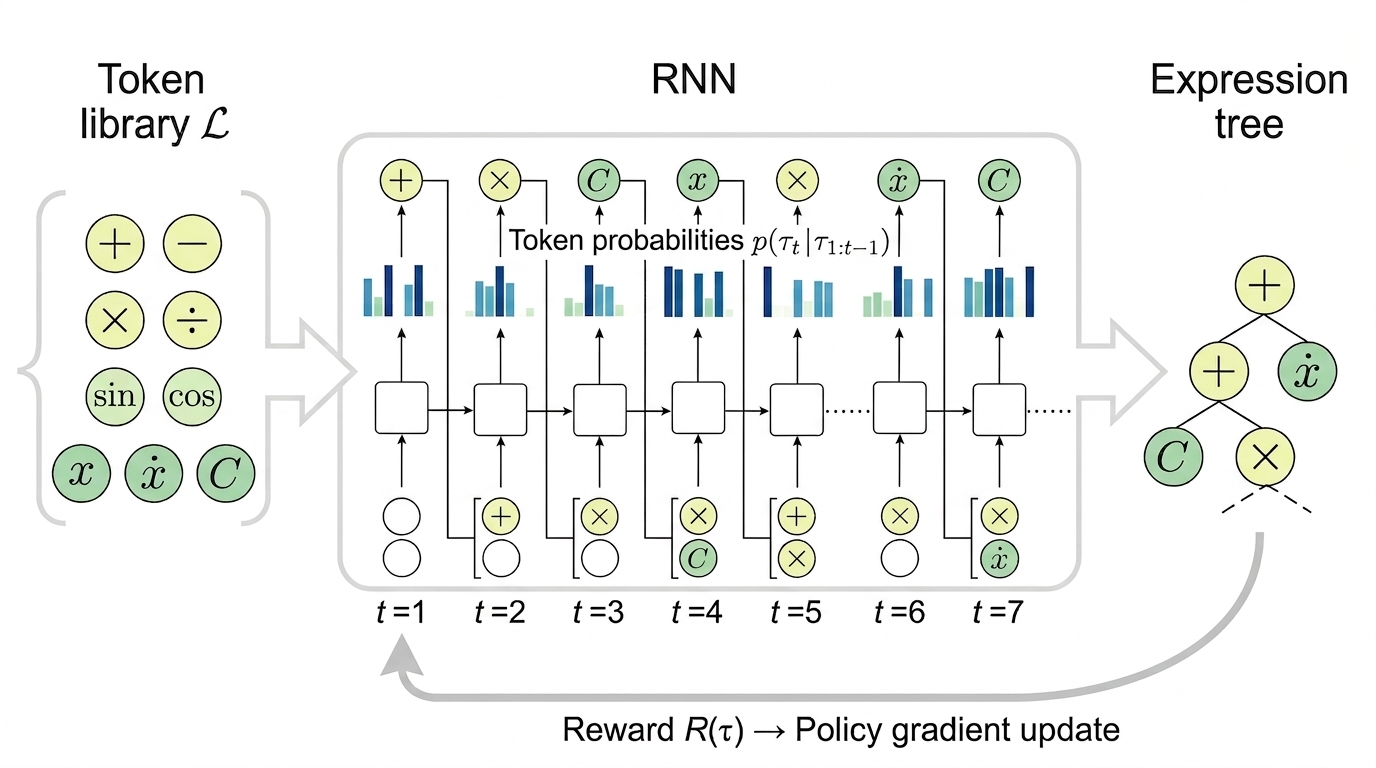}
    \caption{Schematic overview of Deep Symbolic Regression (DSR). A recurrent neural network (RNN) generates symbolic expressions token by token, building expression trees from a library of primitives. Each expression is evaluated on the input data and assigned a reward; the RNN is updated via a risk-seeking policy gradient to favour the highest-reward expressions.}
    \label{fig:dsr_schematic}
\end{figure}

\subsection*{Expression generation via a recurrent neural network}
DSR represents mathematical expressions as sequences of tokens drawn from a user-specified library $\mathcal{A}$.
This library defines the building blocks available to the algorithm and typically includes arithmetic operators (e.g., $+, -, \times, \div$), elementary functions (e.g., $\sin, \cos, \exp, \log$), state variables ($\left\{x_1, \ldots, x_n\right\}$ and inputs $u$),
and ephemeral constants that are optimized during expression evaluation (see Section~\ref{subsec:const_opt} below).
Each candidate expression is generated one token at a time, conditioned on all previously generated tokens, via a recurrent neural network (RNN), which outputs a probability distribution over tokens at each step. The RNN parameters map to a stochastic policy $\pi$ that assigns a probability to each symbolic expression in the search space.

\subsection*{Training via risk-seeking policy gradient}
The RNN is trained using reinforcement learning. At each training iteration, a batch of $B$ candidate expressions is sampled from the current policy $\pi_\theta$. Each expression $\tau$ is evaluated on the input data to compute a reward $R(\tau)$, typically defined as the negative mean squared error between the expression's predictions and the target values. Rather than optimizing the expected reward over all sampled expressions, DSR employs a \textit{risk-seeking} policy gradient: Only the top-$\epsilon$ fraction of expressions (those with the highest rewards; in our experiments $\epsilon = 5\%$) are used to update the policy. This encourages the RNN to focus on the most promising regions of expression space, accelerating convergence toward high-quality symbolic models.

The policy gradient update takes the form:
\[
\nabla_\theta J \approx \frac{1}{|\mathcal{T}_\epsilon|} \sum_{\tau \in \mathcal{T}_\epsilon} \left( R(\tau) - \bar{R} \right) \nabla_\theta \log \pi_\theta(\tau),
\]
where $\nabla_\theta$ denotes the gradient with respect to the RNN parameters, $\mathcal{T}_\epsilon$ denotes the set comprised of the top-$\epsilon$ sampled expressions, and $\bar{R}$ is the mean reward within this set.

\subsection*{Regularization and constraints}
To promote parsimony and physically meaningful expressions, DSR incorporates several regularization mechanisms:
\begin{itemize}
  \item \textbf{Soft length prior} A penalty that discourages overly long expressions, controlled by a location parameter (preferred length) and a scale parameter (tolerance).
  \item \textbf{Entropy bonus} An entropy regularization term weighted by $w$ and decayed by $\gamma$ over training, which encourages exploration of diverse expressions in early iterations.
  \item \textbf{Constraint enforcement} Domain-specific constraints can be imposed on the generated expressions, such as preventing nested trigonometric functions or enforcing dimensional consistency.
\end{itemize}

\subsection*{Constant optimization}
\label{subsec:const_opt}
Once a symbolic skeleton is generated by the RNN, any ephemeral constants within the expression are optimized using numerical methods (e.g., BFGS) to minimize the fitting error. This decouples the structural search (handled by the RNN) from the parametric optimization, improving both efficiency and accuracy.

\subsection*{Role in our framework}
In our pipeline, DSR is applied to the denoised time series data to discover the functional form of the drift term $\bs{\mu}\left(\bs{X} \right)$. The discovered expression then provides the basis functions used to construct the feature set $\bs{\Theta}$ for the subsequent maximum likelihood estimation of the diffusion term. We emphasize that DSR is one possible choice of symbolic regression backend; the modular design of our framework allows any symbolic regression engine to be substituted without modifying the downstream inference pipeline.

\section{Gaussian process denoising}
\label{app:gpdenoising}
The denoising stage employs Gaussian processes to perform two critical tasks: (i) denoising the raw time series data, and (ii) computing smooth estimates of the state derivatives required for the symbolic regression step.

\subsection*{Kernel choice}
Denoting by $\bs{x} = [x(t_1), \ldots, x(t_m)]^\top$ the observed state trajectory, we model the latent (noise-free) trajectory using a GP with the squared exponential (radial basis function) kernel given by
\[
k(\bs{x}, \bs{x}') = \sigma_f^2 \exp\left( -\frac{\|\bs{x} - \bs{x}'\|^2}{2\ell^2} \right) + \sigma_n^2 \delta(\bs{x}, \bs{x}'),
\]
where $\sigma_f^2$ is the signal variance, $\ell$ is the length scale of the GP prior,
and $\sigma_n^2$ is the observation noise variance. The kernel hyperparameters are optimized by maximizing the marginal likelihood of the observed data~\cite{RasmussenWilliams2006}.

\subsection*{Derivative computation}
A key advantage of GPs is that the derivative of a GP is itself a GP. Given a GP posterior over the state trajectory $\bs{x}(t)$, the derivative $\dot{\bs{x}}(t)$ can be computed analytically through differentiation of the kernel function. For any two time points $t$ and $t'$:
\[
\text{cov}\left[\dot{\bs{x}}(t), \bs{x}(t')\right] = \frac{\partial}{\partial t} k(t, t'), \qquad
\text{cov}\left[\dot{\bs{x}}(t), \dot{\bs{x}}(t')\right] = \frac{\partial^2}{\partial t \, \partial t'} k(t, t').
\]
This yields smooth, robust derivative estimates even in the presence of substantial measurement noise, as demonstrated in \hyperref[fig:framework]{Fig.~\ref*{fig:framework}B} of the main text. In contrast, naive finite difference methods amplify noise and produce highly oscillatory derivative estimates that are unsuitable for symbolic regression.

\subsection*{Computational considerations}
Standard GP regression has $\mathcal{O}(M^3)$ computational complexity due to matrix inversion, where $M$ is the number of time-series observations used for GP regression.
For the dataset sizes considered in this work ($M \sim 10^3$), this cost is manageable. For larger datasets, sparse GP approximations~\cite{RasmussenWilliams2006} could be used without modifying the overall framework.

\subsection*{Derivative estimation for noise extraction}
Derivative estimation serves two distinct purposes, each with different requirements. During \textit{denoising}
(Section~\ref{app:gpdenoising}), the goal is to isolate the deterministic dynamics; therefore, GP-based derivatives are used to suppress noise and obtain smooth estimates suitable for symbolic regression. During \textit{noise extraction}, by contrast, the goal is to retain the full noise content present in the data, as it carries the stochastic signature of the system. For this reason, we employ standard finite differences rather than GP smoothing when computing the residuals used for diffusion estimation. When state derivatives are not directly accessible from the data, this finite difference approximation introduces a small additional source of attenuation in the estimated diffusion amplitudes, as analyzed in Section~\ref{app:em}.

\section{Framework and numerical settings}
\label{app:numerical_settings}
A full description of the DSR hyperparameters can be found in \cite{petersen2021deep} and \cite{landajuela2021improving}.
In our implementation, we disable the optional genetic-programming module of the DSO framework (\verb|gp_meld| = \verb|False|), relying solely on the RNN-based policy for expression generation.
Table~\ref{tab:settings} below summarizes the hyperparameter settings for each problem:
\begin{table}[t]
    \centering
    
    \renewcommand{\arraystretch}{1.5}
    \begin{tabular}{cccccccc}
        \toprule
        & & \multicolumn{2}{c}{Soft Length Prior} &  \multicolumn{2}{c}{Entropy} &&\\
        library & epochs & location & scale & $w$ & $\gamma$ &learning rate &reward\\ \midrule
        
        \multicolumn{1}{l}{\textit{Linear Systems}} &&&&&&& \\ \cmidrule(r){1-2}
        $\{ +,-,\times,C\}$ & $50$ & $4$ & $2$ & $0.02$ & $0.85$ & $5 \times 10^{-4}$ & neg-mse \\ \midrule

        \multicolumn{1}{l}{\textit{Nonlinear Systems}} &&&&&& \\ \cmidrule(r){1-2}
        $\{ +,-,\times,C\}$ & $100$ & $14$ & $2$ & $0.02$ & $0.85$ & $5\times 10^{-4}$ & neg-mse\\
        \midrule

        \multicolumn{2}{l}{\textit{Biological Oscillator}} &&&&&& \\ \cmidrule(r){1-2}
        $\{ +,-,\times,C,\sin\}$ & $50$ & $20$ & $5$ & $0.02$ & $0.85$ & $5 \times 10^{-4}$ & neg-mse\\
        \bottomrule
    \end{tabular}
    \renewcommand{\arraystretch}{1}
\caption{Hyperparameter settings used for each of the problems studied.}
\label{tab:settings}
\end{table}

\noindent In the table, $C$ denotes an ephemeral constant whose value is optimized via BFGS during expression evaluation. \textsf{location} and \textsf{scale} are the parameters of the soft length prior, controlling the preferred expression length and tolerance, respectively. $w$ is the entropy regularization weight and $\gamma$ its decay factor per epoch. \textsf{neg-mse} denotes the negative mean squared error, used as the reward signal for the reinforcement learning update.
\noindent For the settings of the Gaussian processes, we set the optimizer to restart $10$ times to reduce the probability of converging to an unfavorable local minimum of the marginal likelihood.
Finally, the automatic relevance determination (ARD) threshold was set to $10^{-8}$ to filter out statistically insignificant variances inferred in the diffusion term by our framework.

\noindent To more accurately control the amount of noise introduced into the simulated systems, we adopt a method that references the magnitude of a state variable rather than directly setting noise as a percentage of the corresponding parameter. For instance, consider the scenario where we aim to introduce a noise level of $5\%$ to the parameter of the forcing factor $\cos(\Omega t)$. In this context, the standard deviation  $\sigma_f$ is calculated as
\begin{equation*}
    \sigma_f = \left( \frac{\sum_{i=1}^{M} \|x_i\|_2 / M}{\sum_{i=1}^{M} \|u_i\|_2 / M} \right) \cdot 0.05,
\end{equation*}
where $M$ is the number of data points (the same $M$ defined in Section~\ref{app:gpdenoising}). This formulation accounts for the ratio of the magnitudes of the reference quantity, in this case $x$, to the target quantity $u$. As an illustrative example, consider the linear system
\[
\ddot{x} = -x - 0.2\dot{x} + 0.4\cos(\Omega t).
\]
Suppose we simulate this deterministic system for 50 seconds using the Euler--Maruyama scheme. The previous calculation then simplifies to
\begin{equation*}
    \sigma_f = \left( \frac{\sim 1}{\sim 0.66} \right) \cdot 0.05 = 0.075.
\end{equation*}
The diffusion---assuming noise is present only in the forcing factor---then becomes
\begin{equation*}
    \sigma = 0.075 \cos(\Omega t).
\end{equation*}

\section{Additional results}
\label{app:additionalresults}

\subsection{Cases of inaccurate symbolic regression results}
\begin{figure}[t]
    \centering
    \includegraphics[width=0.9\textwidth]{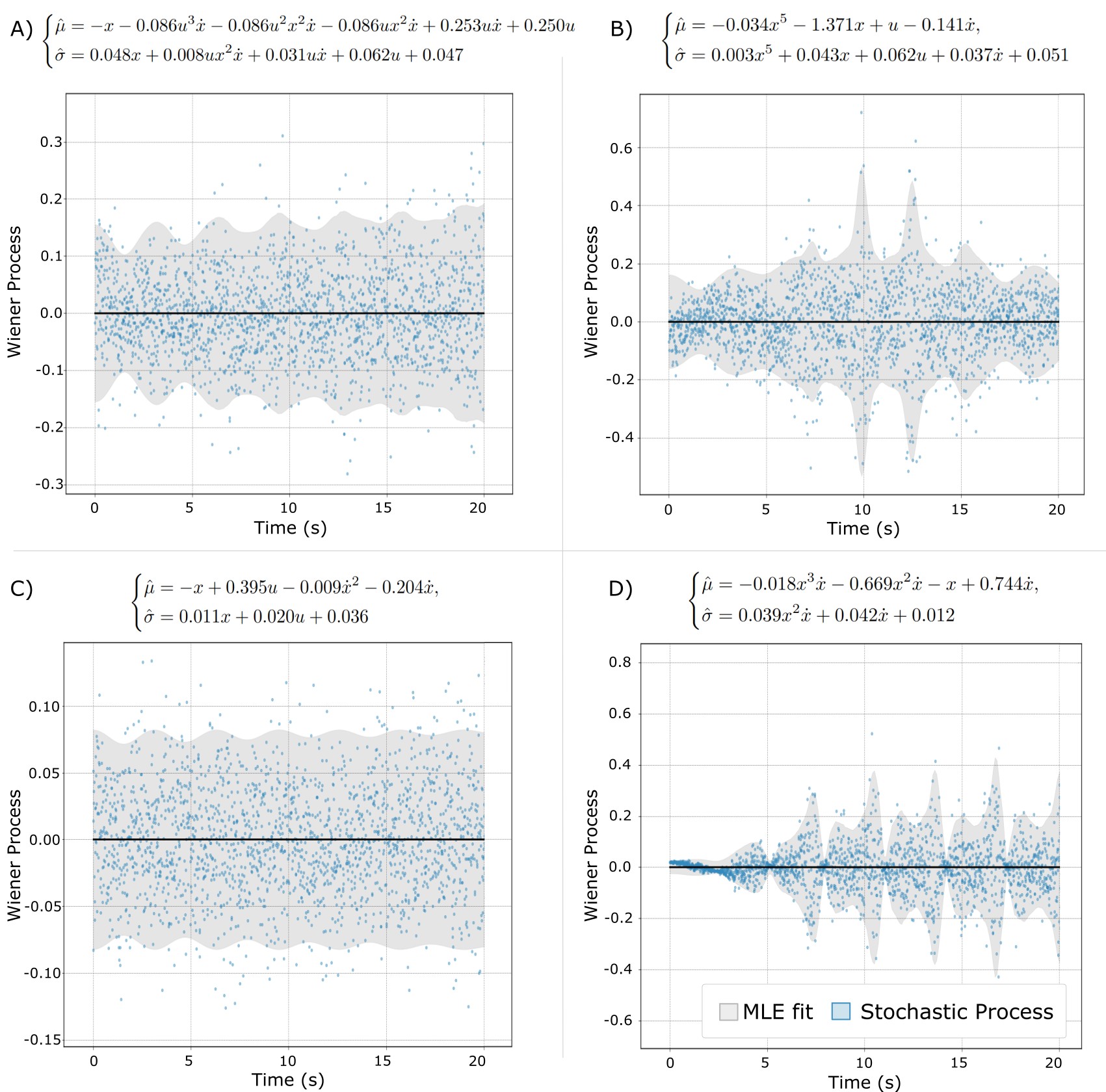}
        \caption{Results from applying the GP-DSR framework to four benchmark SDEs under broad DSR regularization settings (soft length-prior location: 20 tokens, scale: 10 tokens; noise level: 5\%). The equations at the top of each panel are the DSR-predicted drift $\hat{\mu}$ and diffusion $\hat{\sigma}$; the black curve shows the GP-predicted mean trajectory; the blue dots are the extracted stochastic residuals; and the grey band is the GP posterior uncertainty. No reference trajectory is superimposed because the predicted symbolic form differs structurally from the true equation. \textbf{(A)} Linear oscillator with additive noise; \textbf{(B)} Duffing oscillator with noise affecting all parameters; \textbf{(C)} Linear oscillator with noise affecting all parameters; \textbf{(D)} Van der Pol oscillator with noise on both the nonlinear and linear damping terms.}
    \label{fig:add_res}
\end{figure}
In this section, we investigate instances where symbolic regression fails to recover the correct mathematical expressions. Selecting appropriate regularization parameters is challenging in practice, especially when the structure of the underlying system is not known in advance. To illustrate the behavior of our framework under such conditions, we deliberately use broad DSR settings: the soft length-penalty prior---which penalizes excessively long symbolic expressions---is parameterized with a location of 20 tokens (the preferred expression length) and a scale of 10 tokens (the tolerance around that preferred length). This broad prior allows the DSR to explore a wide range of expression lengths without strong regularization pressure, simulating a setting where no prior knowledge of the expression complexity is available. We conduct tests at a noise level of \(5\%\), with DSR training capped at 200 epochs.

Figure~\ref{fig:add_res} shows cases where the DSR component of our framework failed to recover the correct symbolic form of the SDE. In each panel, the equations displayed at the top are the expressions returned by DSR (i.e., the predicted drift $\hat{\mu}$ and diffusion $\hat{\sigma}$), not the reference equations. No reference curves are shown because the true drift differs structurally from the identified expression; plotting the reference trajectory alongside a qualitatively wrong model would be misleading. The black curve is the predicted mean trajectory $\hat{\mu}$ obtained from the GP fit, while the blue dots are the residuals (the stochastic component extracted after subtracting $\hat{\mu}$ from the data), and the grey band represents the GP posterior uncertainty. Noteworthy, although DSR did not recover the correct symbolic expressions in these cases, the GP component still captures the dominant drift well enough that the predicted mean remains a reasonable description of the data. This illustrates that the GP fit is relatively robust to symbolic regression failures, provided the drift model is structurally flexible enough to approximate the true dynamics.

\section{Derivation of the noisy Adler equation}
\label{app:adler}
As discussed by Japaridze et al.~\cite{japaridze2024synchronization}, the phase dynamics of each bacterial oscillator in the coupled cavity system follow
\begin{align*}
  \dv{\phi_1}{t} &= \omega_1 + \frac{K}{2} \sin(\phi_2 - \phi_1) - \frac{\delta v_1}{r}, \\
  \dv{\phi_2}{t} &= \omega_2 + \frac{K}{2} \sin(\phi_1 - \phi_2) - \frac{\delta v_2}{r},
\end{align*}
where $\phi_1$ and $\phi_2$ are the instantaneous phases of the two oscillators, $\omega_1$ and $\omega_2$ are their respective natural (intrinsic) frequencies, and $K$ is the coupling strength between them. The noise terms $\delta v_1$ and $\delta v_2$ represent the independent velocity fluctuations of each bacterium within the shared cavity; with $r$ denoting the cavity radius, the factor $1/r$ converts the velocity fluctuations into phase fluctuations. The fluctuations $\delta v_i$ are modeled as zero-mean, delta-correlated Gaussian noise satisfying $\langle \delta v_i(t)\,\delta v_i(t+\tau) \rangle = 2\sigma_v^2\,\delta_{\mathrm{D}}(\tau)$ (with $\langle \delta v_1(t),\,\delta v_2(t')\rangle = 0$), where $\langle\cdot\rangle$ denotes the ensemble average, $\delta_{\mathrm{D}}$ the Dirac delta distribution, and $\sigma_v$ the noise amplitude.

Defining the phase difference $\varphi = \phi_2 - \phi_1$ and the frequency mismatch $\Delta\omega = \omega_2 - \omega_1$, and subtracting the equations above, we obtain the noisy Adler equation
\begin{equation*}
  \dv{\varphi}{t} = \Delta\omega - K \sin(\varphi) + \xi(t),
\end{equation*}
where $\xi(t)$ is the effective additive noise in the phase-difference equation. In the original model, each oscillator is subject to independent velocity fluctuations $\delta v_1(t)$ and $\delta v_2(t)$, each delta-correlated with intensity $\sigma_v^2$. Upon subtraction, the noise term becomes $\xi(t) = -(1/r)\,\delta v_1(t) + (1/r)\,\delta v_2(t)$, which is a zero-mean Gaussian process with effective diffusion amplitude $(\sigma_v/r)\sqrt{2}$. The composite noise $\xi(t)$ inherits the delta-correlation property and enters the Adler equation as a single additive noise term.

\section{Comparison with BISDEs on synchronization data}
\label{app:bact}
For the experiment with the Bayesian identification of SDEs (BISDEs) algorithm~\cite{WANG2022244}, we constructed the token library to partially cover the space of expressions that our DSR framework explores, so that the difficulty of the identification problem is comparable for both methods. Using $\phi_1$ and $\phi_2$ to denote the two oscillator phases (consistent with the notation of Section~\ref{app:adler}), the token library is:
\begin{equation*}
  \Bigl\{1,\, \sin(\phi_i),\, \sin(\phi_i^2),\,\phi_i\sin(\phi_j),\, \sin\!\bigl(k(\phi_1-\phi_2)\bigr),\, \sin(\phi_1+\phi_2) \Bigr\}, \qquad i,j\in\{1,2\},\; k=\left\{1,2,3\right\}.
\end{equation*}

Additionally, we set the \verb|prune-gamma| parameter (analogous to our ARD threshold, but applied during training rather than as a post-processing step) to $10^{-3}$, as recommended by Wang et al.~\cite{WANG2022244}. Table~\ref{tab:rb} compares the results of GP-DSR and BISDEs on the two synchronization datasets against the empirical method of Japaridze et al.~\cite{japaridze2024synchronization}.

\begin{table}[t]
\renewcommand{\arraystretch}{1.3}
\begin{tabularx}{\textwidth}{lX}
\toprule
\multicolumn{2}{c}{\textbf{GP-DSR}} \\
\midrule
\multirow{2}{*}{\textit{Test 1}}  & $\displaystyle \dd{\phi_1} = \left(0.211 \sin\left(\phi_1 - 0.993 \phi_2\right) - 2.507\right)\,\dd{t} + \left(0.093 \sin\left(\phi_1 - 0.993 \phi_2\right) + 0.352\right)\,\dd{W}$ \\
                          & $\displaystyle \dd{\phi_2} = \left(0.107 \sin\left(0.802 \phi_1 - \phi_2\right) - 2.790\right)\,\dd{t} + \left(0.101 \sin\left(0.802 \phi_1 - \phi_2\right) + 0.258\right)\,\dd{W}$                                                                                                                                                                           \\[2pt] \coloredhline{gray!40} \\[-12pt]
\multirow{2}{*}{\textit{Test 2}}  & $\displaystyle \dd{\phi_1} = \left(0.108 \sin\left(1.830 \phi_1 - 2 \phi_2\right) - 3.057\right)\,\dd{t} + \left(0.270 \sin\left(1.830\phi_1 - 2 \phi_2\right) + 0.316\right)\,\dd{W}$ \\
                          & $\displaystyle \dd{\phi_2} = \left(0.163 \sin\left(0.932 \phi_1 - 0.965\phi_2\right) - 2.790\right)\,\dd{t} + 0.211\,\dd{W}$ \\ \midrule
\multicolumn{2}{c}{\textbf{BISDEs}} \\
\midrule
\multirow{2}{*}{\textit{Test 1}} & $ \begin{aligned} \displaystyle \dd{\phi_1} &= \left(0.1564 \sin(\phi_1) + 0.1335 \sin(\phi_1 + \phi_2) - 0.0727 \sin(3(\phi_1 - \phi_2)) - 2.5709\right)\,\dd{t} \\ & \quad  +\sqrt{0.0519 \sin(\phi_1) + 0.1208}\,\dd{W} \end{aligned}$ \\
                          & $\begin{aligned} \displaystyle \dd{\phi_2} & = \left(-0.0371 \sin(\phi_2) + 0.0395 \sin(\phi_1 + \phi_2) + 0.0595 \sin(\phi_1 - \phi_2) - 0.0400 \sin(\phi_1^2) - 2.7544\right)\,\dd{t} \\ & \quad  + 0.265 \, \dd{W} \end{aligned}$ \\[2pt] \coloredhline{gray!40} \\[-12pt]                        
\multirow{2}{*}{\textit{Test 2}} & $\begin{aligned} \displaystyle \dd{\phi_1} &= \left(-0.079\sin\left(2\left(\phi_1 - \phi_2\right)\right) + 3.103\right)\,\dd{t} \\
 & \quad + \sqrt{0.0571 \sin\left(\phi_1\right) + 0.0743 \sin\left(\phi_2^2\right) + 0.0533 \sin\left(\phi_1 + \phi_2\right) + 0.1260}\,\dd{W}\end{aligned} $ \\
                          & $\begin{aligned} \displaystyle \dd{\phi_2} &= \bigl(-0.0714 \sin\left(\phi_2 \right) + 0.0439 \sin\left(3\left(\phi_1 - \phi_2\right)\right) - 0.0423 \sin\left(2\left(\phi_1 - \phi_2\right)\right) \\ & \qquad + 0.0231 \sin\left(\phi_1 - \phi_2\right) + 0.0683 \sin\left(\phi_1 + \phi_2\right) + 3.0847\bigr)\,\dd{t} + 0.209 \, \dd{W} \end{aligned}$ \\ \midrule
\multicolumn{2}{c}{\textbf{Empirical Method}} \\ \midrule
\multirow{2}{*}{\textit{Test 1}} & $\displaystyle \dd{\phi_1} = \left(0.130 \sin\left(\phi_1 - \phi_2\right) - 2.550\right)\,\dd{t} + 0.300 \, \dd{W}$                                                                                                                                                                   \\
                          & $\displaystyle \dd{\phi_2} = \left(0.130 \sin\left(\phi_1 - \phi_2\right) - 2.800\right)\,\dd{t} + 0.300 \, \dd{W}$                                                                                                                                                                   \\[2pt] \coloredhline{gray!40} \\[-12pt]
\multirow{2}{*}{\textit{Test 2}} & $\displaystyle \dd{\phi_1} = \left(0.085 \sin\left(\phi_1 - \phi_2\right) + 3.100\right)\,\dd{t} + 0.300 \, \dd{W}$                                                                                                                                                                   \\
                          & $\displaystyle \dd{\phi_2} = \left(0.085 \sin\left(\phi_1 - \phi_2\right) + 3.040\right)\,\dd{t} + 0.300 \, \dd{W}$  \\ \bottomrule
\end{tabularx}
\caption{Equations for GP-DSR, BISDEs, and Empirical Method}
\label{tab:rb}
\end{table}

\noindent It is worth noting that the diffusion processes in BISDEs are represented using a \emph{square root} because BISDEs adopt a slightly different stochastic differential equation formulation. Specifically, they regress the \emph{symbolic form of a single diffusion process}, corresponding to the \emph{variance}, leading to an equation of the form
\begin{equation}
    \dd{\phi(t)} = f(\phi)\, \dd{t} + \sqrt{g(\phi)}\, \dd{W(t)},
\end{equation}
where $f(\phi)$ is the drift function, $g(\phi)$ is the diffusion variance, and $W(t)$ is a scalar Wiener process.
In contrast, our approach allows for \emph{multiple diffusion processes}, each modeled \emph{directly as a standard deviation function} (i.e., without the square root):
\begin{equation}
    \dd{\boldsymbol{\phi}(t)} = f(\boldsymbol{\phi})\, \dd{t} + \mathbf{G}(\boldsymbol{\phi})\, \dd{\mathbf{W}(t)},
\end{equation}
where $f(\boldsymbol{\phi})$ is the drift vector, $\mathbf{G}(\boldsymbol{\phi})$ is the diffusion matrix whose columns are standard deviation functions, and $\mathbf{W}(t)$ is a vector-valued Wiener process.
\noindent In our reported results, we write all diffusion processes using a single symbol $\dd{W}$ only for notational simplicity, even though each diffusion term is modeled as an independent stochastic process.

\section{Detailed numerical results}
\label{app:resultstats}

\begin{figure}[t]
    \centering
    \includegraphics[width=\textwidth]{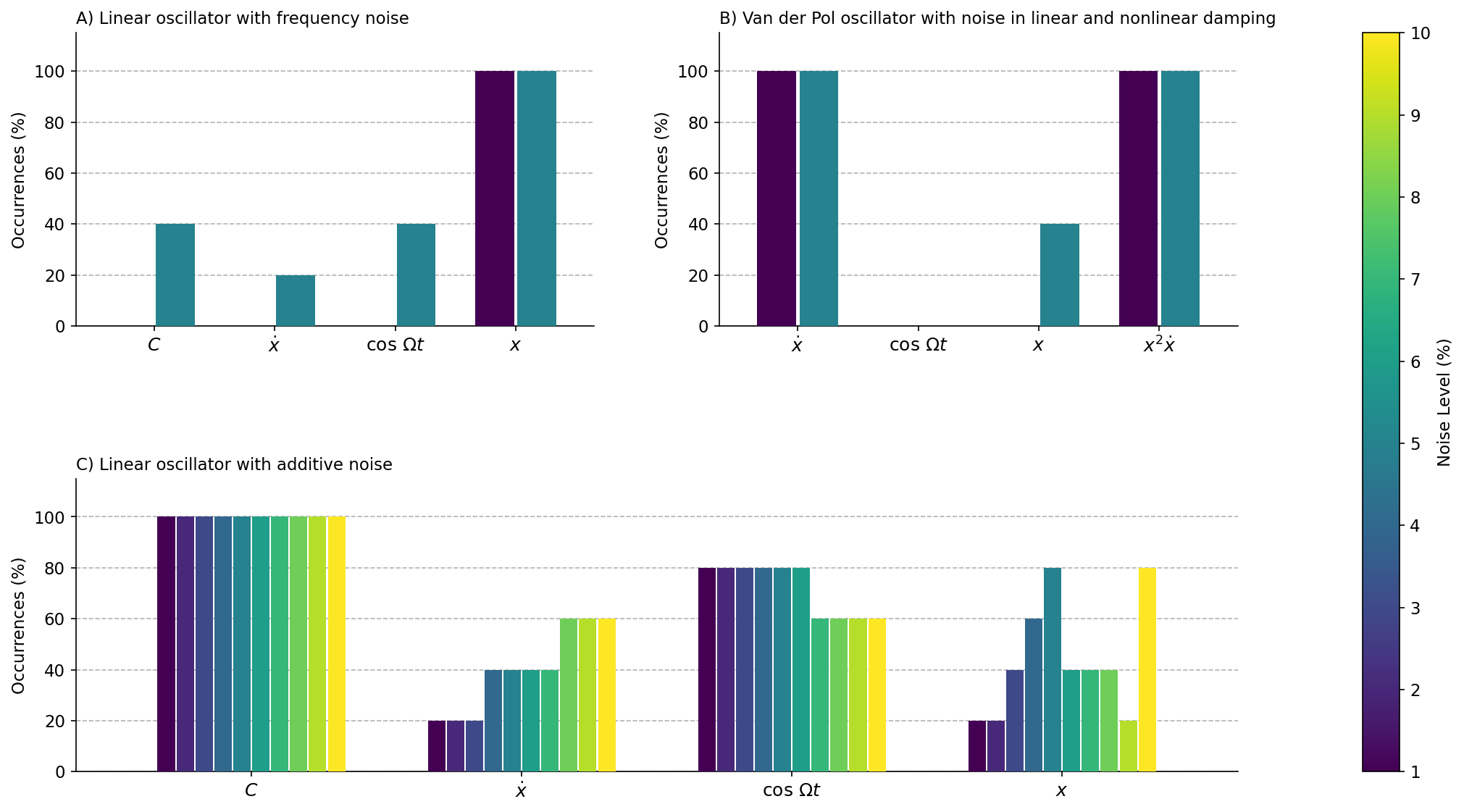}
    \caption{
        Bar plots showing the occurrence frequency of each diffusion token in the model predictions, for various systems and noise levels. (\textbf{A}) Linear oscillator with frequency noise, (\textbf{B}) Van der Pol oscillator with noise in linear and nonlinear damping, and (\textbf{C}) Linear oscillator with additive noise. Each bar is computed from 5 independent realizations for each noise level. The framework consistently captures the correct tokens across all realizations; minor spurious activations are observed in some realizations at higher noise levels.
    }
    \label{fig:token_occurrences}
\end{figure}

\paragraph{Summary of key trends} In all tables below, $\Omega$ denotes the forcing frequency of the periodic driving term (defined in Section~\ref{app:euler}).
Across all systems, drift coefficients remain close to ground truth throughout the entire noise range tested ($1$--$10\%$ for the additive case; $1\%$ and $5\%$ for the remaining cases). At $1\%$ noise, drift estimates typically fall within $1$--$3\%$ of the true values; at $5\%$ noise, deviations remain below $5$--$12\%$ for most coefficients, with uncertainty intervals widening gradually with noise. Diffusion amplitudes are systematically underestimated by approximately $6$--$20\%$ relative to the ground truth---an effect arising from the Euler--Maruyama approximation used in noise extraction and analyzed in detail in Section~\ref{app:em}. In all tables below, each predicted coefficient is reported as $\text{mean}_{\pm\,\text{std}}$, where mean and standard deviation are computed over five independent experimental realizations. The ARD pruning consistently suppresses basis functions absent from the true diffusion; however, small spurious activations appear in the diffusion predictions at all noise levels, and their magnitude grows with increasing noise intensity (see Figure~\ref{fig:token_occurrences} for a summary of token occurrence frequencies). These spurious contributions remain small relative to the active terms and are a known consequence of the Euler--Maruyama attenuation bias discussed in the main text.

\begin{table}[t]
    \centering
    \renewcommand{\arraystretch}{2.0}
    \begin{tabular}{p{0.15\textwidth} p{0.8\textwidth}}
    \toprule
        \multicolumn{2}{l}{\textbf{Linear oscillator with additive noise}} \\
        \midrule
        \textit{Drift}: & $\mu = -x + 0.4\cos(\Omega t) - 0.2\dot{x}$ \\
        \textit{Diffusion}: & $\sigma = \sigma_a$ \quad (additive noise; $\sigma_a = 0.01 \times p$ where $p$ is the noise level in percent, e.g., $\sigma_a = 0.01$ at $1\%$, $\sigma_a = 0.1$ at $10\%$) \\
        \midrule
        \textit{Noise Level} & \textit{Model Predictions} \\
        1\% & 
        \(\begin{aligned}
        \hat{\mu} &= -\underset{\pm 0.0111}{1.0052}x + \underset{\pm 0.0330}{0.4031}\cos(\Omega t) - \underset{\pm 0.0216}{0.2098}\dot{x} \\
        \hat{\sigma} &= \underset{\pm 0.0009}{0.0011}x + \underset{\pm 0.0018}{0.0030}\cos(\Omega t) + \underset{\pm 0.0007}{0.0094}
        \end{aligned}\) \\[12pt] \coloredhline{gray!40} \\[-16pt]
        2\% & 
        \(\begin{aligned}
        \hat{\mu} &= -\underset{\pm 0.0093}{0.9978}x + \underset{\pm 0.0412}{0.3947}\cos(\Omega t) - \underset{\pm 0.0335}{0.2039}\dot{x} \\
        \hat{\sigma} &= \underset{\pm 0.0028}{0.0045}\cos(\Omega t) + \underset{\pm 0.0005}{0.0003}\dot{x} + \underset{\pm 0.0008}{0.0201}
        \end{aligned}\) \\[12pt] \coloredhline{gray!40} \\[-16pt]
        3\% & 
        \(\begin{aligned}
        \hat{\mu} &= -\underset{\pm 0.0100}{0.9988}x + \underset{\pm 0.0317}{0.4140}\cos(\Omega t) - \underset{\pm 0.0119}{0.2170}\dot{x} \\
        \hat{\sigma} &= \underset{\pm 0.0036}{0.0018}x + \underset{\pm 0.0070}{0.0096}\cos(\Omega t) + \underset{\pm 0.0012}{0.0009}\dot{x} + \underset{\pm 0.0040}{0.0291}
        \end{aligned}\) \\[12pt] \coloredhline{gray!40} \\[-16pt]
        4\% & 
        \(\begin{aligned}
        \hat{\mu} &= -\underset{\pm 0.0346}{1.0082}x + \underset{\pm 0.0819}{0.3871}\cos(\Omega t) - \underset{\pm 0.0439}{0.2030}\dot{x} \\
        \hat{\sigma} &= \underset{\pm 0.0068}{0.0053}x + \underset{\pm 0.0108}{0.0159}\cos(\Omega t) + \underset{\pm 0.0016}{0.0012}\dot{x} + \underset{\pm 0.0083}{0.0373}
        \end{aligned}\) \\[12pt] \coloredhline{gray!40} \\[-16pt]
        5\% & 
        \(\begin{aligned}
        \hat{\mu} &= -\underset{\pm 0.0273}{0.9932}x + \underset{\pm 0.0746}{0.3842}\cos(\Omega t) - \underset{\pm 0.0400}{0.2011}\dot{x} \\
        \hat{\sigma} &= \underset{\pm 0.0060}{0.0030}x + \underset{\pm 0.0116}{0.0153}\cos(\Omega t) + \underset{\pm 0.0021}{0.0015}\dot{x} + \underset{\pm 0.0069}{0.0491}
        \end{aligned}\) \\[12pt] \coloredhline{gray!40} \\[-16pt]
        6\% & 
        \(\begin{aligned}
        \hat{\mu} &= -\underset{\pm 0.0326}{0.9839}x + \underset{\pm 0.1053}{0.3568}\cos(\Omega t) - \underset{\pm 0.0528}{0.1901}\dot{x} \\
        \hat{\sigma} &= \underset{\pm 0.0029}{0.0015}x + \underset{\pm 0.0087}{0.0098}\cos(\Omega t) + \underset{\pm 0.0051}{0.0028}\dot{x} + \underset{\pm 0.0052}{0.0597}
        \end{aligned}\) \\[12pt] \coloredhline{gray!40} \\[-16pt]
        7\% & 
        \(\begin{aligned}
        \hat{\mu} &= -\underset{\pm 0.0453}{1.0090}x + \underset{\pm 0.0682}{0.3903}\cos(\Omega t) - \underset{\pm 0.0232}{0.1983}\dot{x} \\
        \hat{\sigma} &= \underset{\pm 0.0104}{0.0086}x + \underset{\pm 0.0165}{0.0183}\cos(\Omega t) + \underset{\pm 0.0087}{0.0070}\dot{x} + \underset{\pm 0.0101}{0.0670}
        \end{aligned}\) \\[12pt] \coloredhline{gray!40} \\[-16pt]
        8\% & 
        \(\begin{aligned}
        \hat{\mu} &= -\underset{\pm 0.0590}{1.0041}x + \underset{\pm 0.0788}{0.3778}\cos(\Omega t) - \underset{\pm 0.0392}{0.1848}\dot{x} \\
        \hat{\sigma} &= \underset{\pm 0.0108}{0.0085}x + \underset{\pm 0.0110}{0.0092}\cos(\Omega t) + \underset{\pm 0.0100}{0.0068}\dot{x} + \underset{\pm 0.0092}{0.0774}
        \end{aligned}\) \\[12pt] \coloredhline{gray!40} \\[-16pt]
        9\% & 
        \(\begin{aligned}
        \hat{\mu} &= -\underset{\pm 0.0919}{1.0329}x + \underset{\pm 0.0736}{0.3771}\cos(\Omega t) - \underset{\pm 0.0556}{0.1801}\dot{x} \\
        \hat{\sigma} &= \underset{\pm 0.0114}{0.0196}x + \underset{\pm 0.0064}{0.0032}\cos(\Omega t) + \underset{\pm 0.0115}{0.0066}\dot{x} + \underset{\pm 0.0076}{0.0842}
        \end{aligned}\) \\[12pt] \coloredhline{gray!40} \\[-16pt]
        10\% & 
        \(\begin{aligned}
        \hat{\mu} &= -\underset{\pm 0.0902}{1.0155}x + \underset{\pm 0.1575}{0.4386}\cos(\Omega t) - \underset{\pm 0.1287}{0.2059}\dot{x} \\
        \hat{\sigma} &= \underset{\pm 0.0112}{0.0175}x + \underset{\pm 0.0205}{0.0257}\cos(\Omega t) + \underset{\pm 0.0084}{0.0089}\dot{x} + \underset{\pm 0.0166}{0.0858}
        \end{aligned}\) \\
    \bottomrule
    \end{tabular}
\end{table}

\begin{table}[t]
    \centering
    \renewcommand{\arraystretch}{2.0}
    \begin{tabular}{p{0.15\textwidth} p{0.8\textwidth}}
    \toprule
        \multicolumn{2}{l}{\textbf{Linear oscillator with frequency noise}} \\
        \midrule
        \textit{Drift}: & $\mu = -x + 0.4\cos(\Omega t) - 0.2\dot{x}$ \\
        \textit{Diffusion}: & $\sigma = \sigma_{w}x$ \quad ($\sigma_w = 2\omega_0 \Delta\omega_0$ is the frequency-noise amplitude; see main text) \\
        \multicolumn{2}{l}{\footnotesize Results are shown at $1\%$ and $5\%$ noise, representative of the range tested for all multiplicative-noise cases.} \\
        \midrule
        \textit{Noise Level} & \textit{Model Predictions} \\
        1\% &
        \(\begin{aligned}
        \hat{\mu} &= -\underset{\pm 0.0008}{1.0005}x + \underset{\pm 0.0007}{0.3998}\cos(\Omega t) - \underset{\pm 0.0031}{0.1984}\dot{x} \\
        \hat{\sigma} &= \underset{\pm 0.0005}{0.0102}x
        \end{aligned}\) \\
        5\% &
        \(\begin{aligned}
        \hat{\mu} &= -\underset{\pm 0.0152}{1.0012}x + \underset{\pm 0.0228}{0.3754}\cos(\Omega t) - \underset{\pm 0.0154}{0.1802}\dot{x} \\
        \hat{\sigma} &= \underset{\pm 0.0016}{0.0504}x + \underset{\pm 0.0011}{0.0008}\cos(\Omega t) + \underset{\pm 0.0005}{0.0002}\dot{x} + \underset{\pm 0.0010}{0.0008}
        \end{aligned}\) \\
        \multicolumn{2}{l}{\footnotesize\textit{Note:} At $5\%$ noise, the spurious $\cos(\Omega t)$, $\dot{x}$, and constant terms in $\hat{\sigma}$ are small and are artifactual activations consistent with the Euler--Maruyama bias.} \\
    \midrule
        \multicolumn{2}{l}{\textbf{Duffing oscillator with noise in every parameter}} \\
        \midrule
        \textit{Drift}: & $\mu = -0.2x^3 - x + 0.7\cos(\Omega t) - 0.1\dot{x}$ \\
        \textit{Diffusion}: & $\sigma = \sigma_{k_3}x^3 + \sigma_w x + \sigma_f\cos(\Omega t) + \sigma_d\dot{x} + \sigma_a$
        \hspace{1em}\footnotesize($\sigma_{k_3}$: cubic-stiffness noise; $\sigma_w$: frequency noise; $\sigma_f$: forcing noise; $\sigma_d$: damping noise; $\sigma_a$: additive noise; see Section~\ref{app:euler}) \\
        \multicolumn{2}{l}{\footnotesize Results shown at $1\%$ and $5\%$ noise, representative of the range tested for all multiplicative-noise cases.} \\
        \midrule
        \textit{Noise Level} & \textit{Model Predictions} \\
        1\% & 
        \(\begin{aligned}
        \hat{\mu} &= -\underset{\pm 0.0125}{0.2084}x^3 - \underset{\pm 0.0314}{0.9621}x + \underset{\pm 0.0105}{0.6752}\cos(\Omega t) - \underset{\pm 0.0028}{0.1064}\dot{x} \\
        \hat{\sigma} &= \underset{\pm 0.0009}{0.0055}x^3 + \underset{\pm 0.0012}{0.0091}x + \underset{\pm 0.0031}{0.0162}\cos(\Omega t) + \underset{\pm 0.0014}{0.0095}\dot{x} + \underset{\pm 0.0008}{0.0092}
        \end{aligned}\) \\
        5\% & 
        \(\begin{aligned}
        \hat{\mu} &= -\underset{\pm 0.0450}{0.1780}x^3 - \underset{\pm 0.1120}{1.0450}x + \underset{\pm 0.0480}{0.6220}\cos(\Omega t) - \underset{\pm 0.0120}{0.0880}\dot{x} \\
        \hat{\sigma} &= \underset{\pm 0.0060}{0.0240}x^3 + \underset{\pm 0.0150}{0.0480}x + \underset{\pm 0.0220}{0.0650}\cos(\Omega t) + \underset{\pm 0.0110}{0.0490}\dot{x} + \underset{\pm 0.0120}{0.0460}
        \end{aligned}\) \\
    \midrule
        \multicolumn{2}{l}{\textbf{Van der Pol oscillator with noise in damping terms}} \\
        \midrule
        \textit{Drift}: & $\mu = -0.7x^2\dot{x} - x + 0.2\cos(\Omega t) + 0.7\dot{x}$ \quad (equivalent form: $\alpha(1 - x^2)\dot{x} - x + fu$, with $\alpha = 0.7$, $f = 0.2$, $u = \cos(\Omega t)$) \\
        \textit{Diffusion}: & $\sigma = \sigma_{d_2}x^2\dot{x} + \sigma_d\dot{x}$ \quad ($\sigma_{d_2}$: nonlinear-damping noise; $\sigma_d$: linear-damping noise) \\
        \multicolumn{2}{l}{\footnotesize Results shown at $1\%$ and $5\%$ noise, representative of the range tested for all multiplicative-noise cases.} \\
        \midrule
        \textit{Noise Level} & \textit{Model Predictions} \\
        1\% &
        \(\begin{aligned}
        \hat{\mu} &= -\underset{\pm 0.0185}{0.7024}x^2\dot{x} - \underset{\pm 0.0084}{0.9958}x + \underset{\pm 0.0021}{0.2015}\cos(\Omega t) + \underset{\pm 0.0241}{0.6852}\dot{x} \\
        \hat{\sigma} &= \underset{\pm 0.0008}{0.0102}x^2\dot{x} + \underset{\pm 0.0006}{0.0098}\dot{x}
        \end{aligned}\) \\
        5\% & 
        \(\begin{aligned}
        \hat{\mu} &= -\underset{\pm 0.0520}{0.6920}x^2\dot{x} - \underset{\pm 0.0450}{0.9750}x + \underset{\pm 0.0080}{0.1980}\cos(\Omega t) + \underset{\pm 0.0680}{0.7250}\dot{x} \\
        \hat{\sigma} &= \underset{\pm 0.0050}{0.0515}x^2\dot{x} + \underset{\pm 0.0024}{0.0016}x + \underset{\pm 0.0040}{0.0495}\dot{x} + \underset{\pm 0.0006}{0.0003}
        \end{aligned}\) \\
        \multicolumn{2}{l}{\footnotesize\textit{Note:} At $5\%$ noise, the spurious $x$ and constant terms in $\hat{\sigma}$ are small ($\lesssim 3\%$ of the dominant $x^2\dot{x}$ coefficient) and are consistent with the expected artifactual activations.} \\
    \bottomrule
    \end{tabular}
\end{table}

\section{Stochastic nonlinear system simulation}
\label{app:euler}
We simulated a stochastic nonlinear oscillator in which random forcing is applied exclusively to the highest derivative term. This modeling choice reflects the physical assumption that noise arises from parametric uncertainty in stiffness and damping---sources that act at the force and acceleration level rather than directly at the position or velocity level. The stochastic forcing is modeled as a sum of independent Wiener increments, each scaled by a state-dependent basis function.

The equations were integrated using the Euler--Maruyama method with a fixed time step of $\Delta t_{\mathrm{sim}} = \SI[exponent-mode = scientific, print-unity-mantissa = false]{1e-3}{\second}$ for a total simulated time of $T_{\mathrm{total}} = \SI{50}{\second}$. The resulting data were uniformly downsampled to $\Delta t_{\mathrm{out}} = \SI[exponent-mode = scientific, print-unity-mantissa = false]{1e-2}{\second}$ for further processing, in order to simulate realistic measurement sampling rates, which are typically coarser than the integration time step.

\paragraph{Model definition.} The system state vector is
\[
    \bs{s} = [x, \ \dot{x}, \ y_1, \ y_2]^\top,
\]
where $x$ is the primary displacement coordinate, $\dot{x}$ is the velocity, and $y_1$ and $y_2$ are auxiliary oscillator states used to generate a periodic forcing term \(f\,y_1\). In the absence of nonlinearities and noise, $y_1 \approx \cos(\Omega t)$ and $y_2 \approx \sin(\Omega t)$, where $\Omega$ (denoted $\omega$ in the auxiliary-system equations below; not to be confused with the natural frequencies $\omega_1, \omega_2$ of the bacterial oscillators in Section~\ref{app:adler}) is the driving frequency.

The deterministic dynamics are:
\begin{align*}
  \ddot{x} & = -d \, \dot{x} - \omega_0^2 \, x - k_3 \, x^3 - k_2 \, x^2 - d_2 \, x^2 \dot{x} - d_1 \, x \dot{x} + f \, y_1, \\ 
  \dot{y}_1 & = y_1 - \omega y_2 - y_1 \left(y_1^2 + y_2^2\right), \\ 
  \dot{y}_2 & = y_2 + \omega y_1 - y_2 \left(y_1^2 + y_2^2\right),
\end{align*}
where $d$ is the linear damping coefficient, $\omega_0$ is the natural frequency of the primary oscillator, $k_2$ and $k_3$ are the quadratic and cubic stiffness coefficients, $f$ is the forcing amplitude, $\omega$ is the driving frequency of the auxiliary system (equal to $\Omega$), $d_1$ is the linear state--velocity coupling coefficient, and $d_2$ is the nonlinear (cubic) damping coefficient.

\paragraph{Stochastic term.}
The stochastic perturbation acts only on $\ddot{x}$:
\[
\eta_{\ddot{x}} =
\sigma_d \, \dot{x} \, \xi_d +
\sigma_f \, y_1 \, \xi_f +
\sigma_w \, x \, \xi_w +
\sigma_a \, \xi_a +
\sigma_{k_2} \, x^2 \, \xi_{k_2} +
\sigma_{k_3} \, x^3 \, \xi_{k_3} +
\sigma_{d_1} \, x \dot{x} \, \xi_{d_1} +
\sigma_{d_2} \, x^2 \dot{x} \, \xi_{d_2},
\]
where each $\xi_i \sim \mathcal{N}(0, \Delta t_{\mathrm{sim}})$ denotes an independent Wiener increment; the variance $\Delta t_{\mathrm{sim}}$ reflects the standard Euler--Maruyama discretization of Brownian motion. The $\sigma_{\cdot}$ coefficients control the noise amplitude for their corresponding multiplicative term.

\paragraph{Numerical integration.}
The system is integrated via Euler--Maruyama:
\[
\begin{aligned}
x_{n+1} &= x_n + \dot{x}_n \, \Delta t_{\mathrm{sim}}, \\
\dot{x}_{n+1} &= \dot{x}_n + \left[ \ddot{x}_n + \eta_{\ddot{x},n} \right] \Delta t_{\mathrm{sim}}, \\
y_{1,n+1} &= y_{1,n} + \dot{y}_{1,n} \, \Delta t_{\mathrm{sim}}, \\
y_{2,n+1} &= y_{2,n} + \dot{y}_{2,n} \, \Delta t_{\mathrm{sim}}.
\end{aligned}
\]
After simulation, the trajectory is uniformly downsampled from $\Delta t_{\mathrm{sim}}$ to $\Delta t_{\mathrm{out}} = \SI[exponent-mode = scientific, print-unity-mantissa = false]{1e-2}{\second}$.

\section{Investigation of the Euler--Maruyama approximation for stochastic process estimation}
\label{app:em}
We examine the effects of time step variations on the recovery of stochastic process data using the Euler--Maruyama approximation (Eq.~\eqref{eq:extraction} of the main text). We operate under the assumption that the predicted mean \(\hat{\mu}\) perfectly matches the actual mean \(\mu\), concentrating on the frequency noise within a linear oscillator, analogous to the scenario depicted in Fig.~\ref{fig:lin_w} of the main text. The reference simulation uses a fine integration time step of $\Delta t = \SI[exponent-mode = scientific, print-unity-mantissa = false]{1e-4}{\second}$ via the Euler--Maruyama scheme.

\begin{figure}[t]
    \centering
    \includegraphics[width=\linewidth]{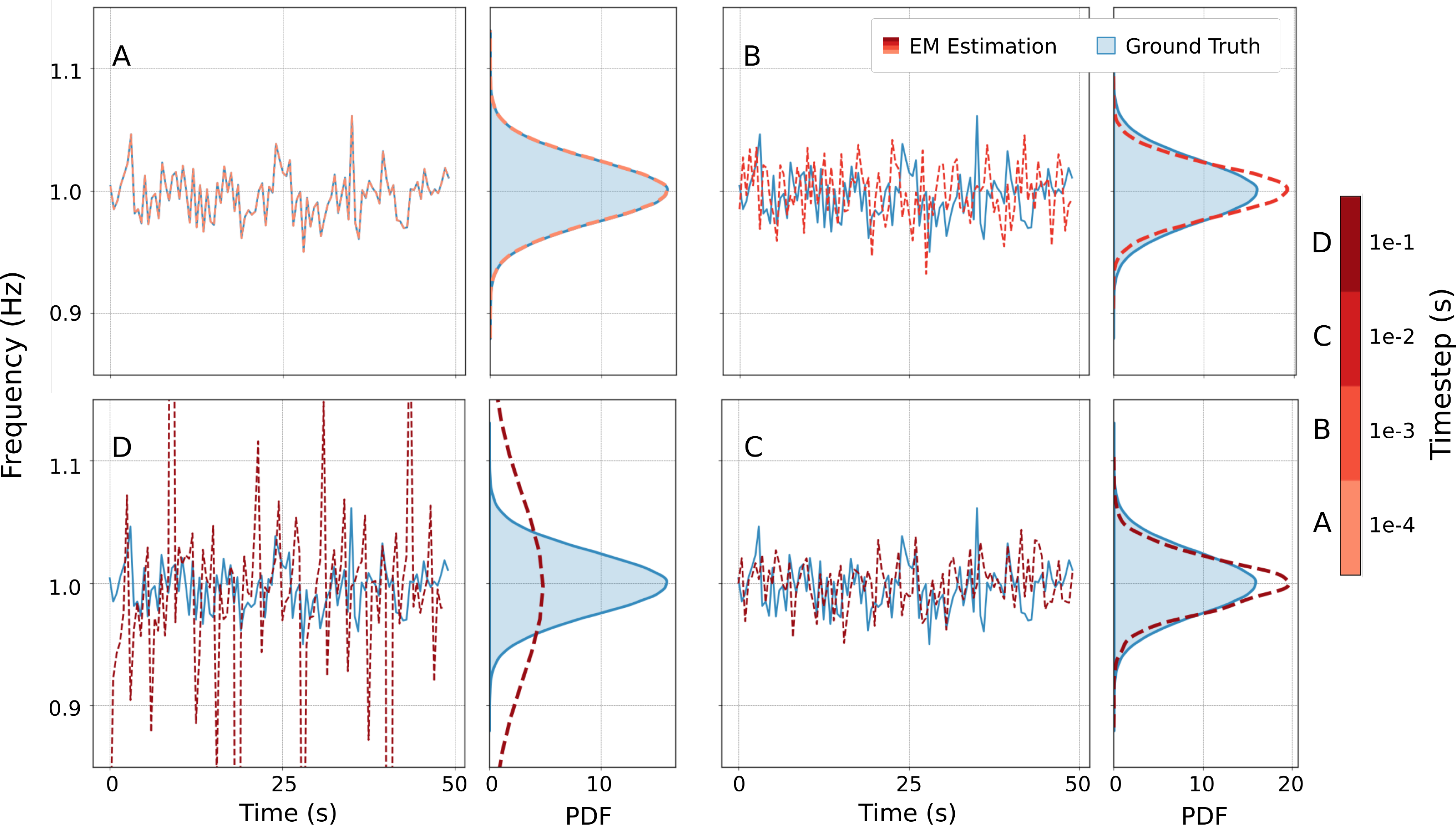}
    \caption{Frequency noise estimation comparison across various time steps \(\Delta t\) (illustrated in red) against the ground truth distribution (in blue).}
    \label{fig:EM-estimation}
\end{figure}

\noindent Figure \ref{fig:EM-estimation} illustrates how the accuracy of the estimation correlates with the selected time step. When the exact discretization time step is employed, the approximation faithfully reconstructs the frequency fluctuations. As the time step increases, however, the estimated distribution broadens and the peak shifts, causing the estimated probability density function to diverge from the ground truth. In practice, measurement data are sampled at intervals much larger than the simulation time step. Using the original integration time step would thus be unrealistic; a larger output time step was chosen to mimic realistic data acquisition rates.

\begin{figure}[t]
    \centering
    \includegraphics[width=\linewidth]{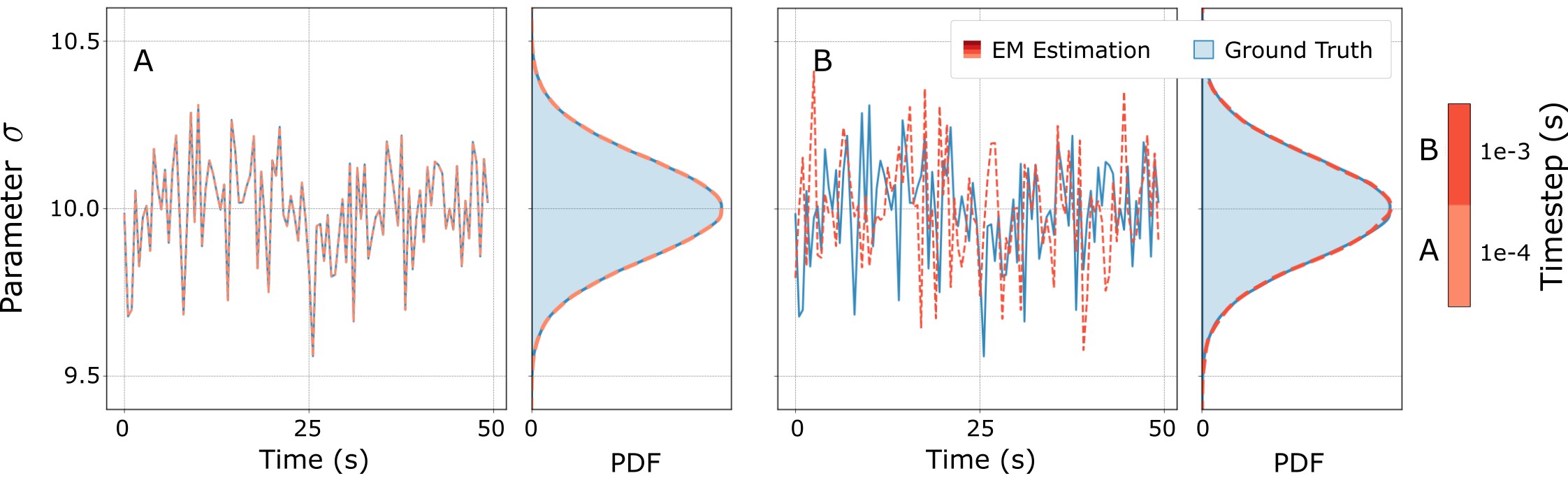}
    \caption{$\sigma_{\mathrm{L}}$ (Lorenz coupling parameter) noise estimation comparison across various time steps \(\Delta t\) (illustrated in red) against the ground truth distribution (in blue).}
    \label{fig:sigma_noise}
\end{figure}

We simulate a Lorenz system to further clarify whether the underestimation bias arises solely from the Euler--Maruyama approximation itself or from the additional error introduced by estimating higher-order derivatives in a second-order differential equation using finite differences. Specifically, we focus on the first Lorenz equation:
\begin{equation*}
  \dot{x} = \sigma_{\mathrm{L}} (y - x),
\end{equation*}
where $\sigma_{\mathrm{L}}$ is the standard Lorenz coupling parameter. We introduce a 5\% noise in the parameter $\sigma_{\mathrm{L}}$ and estimate its distribution \textit{a posteriori} using the same approximation. In this case, the stochastic increment can be estimated directly as $\xi = \dot{x} - \sigma_{\mathrm{L}}(y-x)$ without requiring numerical differentiation. As illustrated in Figure \ref{fig:sigma_noise}, the estimated distribution from the Euler--Maruyama approximation matches the original distribution (shown in blue) even for time steps up to $\Delta t = \SI[exponent-mode = scientific, print-unity-mantissa = false]{1e-2}{\second}$. This indicates that the underestimation bias observed in our experiments is due to the superposition of the Euler--Maruyama approximation, which scales with \(\sqrt{\Delta t}\), and the finite difference method used to estimate state derivatives, which scales with \(\Delta t\), with the former contributing more to the total error than the latter. Both error sources should therefore be considered when selecting the output time step for real-world applications.

\section{Random differential equations vs.\ stochastic differential equations}
\label{app:rode_vs_sde}

Stochastic differential equations (SDEs) and random differential equations (RODEs) are two distinct frameworks for modeling systems subject to randomness, and their difference is relevant to understanding how our framework relates to methods such as HyperSINDy~\cite{jacobs2023hypersindy}.

An SDE in It\^o form reads
\begin{equation*}
    \dd{\bs{X}} = \bs{\mu}(\bs{X})\,\dd{t} + \bs{\sigma}(\bs{X})\,\dd{\bs{W}},
\end{equation*}
where $\bs{\mu}$ is the drift, $\bs{\sigma}$ is the diffusion, and $\dd{\bs{W}}$ is the increment of a Wiener process. The diffusion term is a \emph{separate, explicit} component of the equation, and its functional form $\bs{\sigma}(\bs{X})$ carries physical meaning---it encodes which state-dependent sources of uncertainty drive the system. Recovering $\bs{\sigma}(\bs{X})$ in symbolic form is the primary goal of our framework.

A RODE, by contrast, takes the form
\begin{equation*}
    \dot{\bs{X}} = \bs{f}(\bs{X}, \bs{\xi}(t)),
\end{equation*}
where $\bs{\xi}(t)$ is a stochastic process (not necessarily a Wiener process) that enters directly through the vector field $\bs{f}$. In a RODE, randomness modulates the \emph{coefficients} of the ODE rather than appearing as a separable diffusion term. HyperSINDy adopts this formulation: its learned equations have coefficients driven by a latent random variable, but there is no explicit $\bs{\sigma}(\bs{X})$ to recover.

The two frameworks are mathematically related: under regularity conditions, an SDE driven by a Wiener process can be transformed into a RODE via the Doss--Sussmann transformation~\cite{doss1977liens, sussmann1978gap}. The converse is not generally true---RODEs can be driven by non-Wiener processes (e.g., fractional Brownian motion or colored noise) that do not admit an It\^o representation. For the class of systems considered in this work---where noise arises from parametric uncertainty in a physical model---the SDE formulation is the natural and interpretable choice, as it directly yields the noise amplitude $\bs{\sigma}(\bs{X})$ as a symbolic expression.

\section{Scalability and the curse of dimensionality}
\label{app:curse_of_dimensionality}

Histogram-based regression (HBR)~\cite{Friedrich2011ApproachingCB, friedrich2000extracting} estimates the drift and diffusion coefficients of an SDE by partitioning the state space into a grid of bins and computing local conditional averages within each bin. While conceptually simple, this approach suffers from the \emph{curse of dimensionality}: for a system with state dimension $D$ and $n_{\mathrm{b}}$ bins per dimension, the total number of bins grows as $n_{\mathrm{b}}^D$. Obtaining statistically reliable estimates in each bin therefore requires a number of data points that grows exponentially with $D$, making the method impractical for even moderately high-dimensional systems. In practice, this forces the use of extremely large datasets---on the order of $10^5$ to $10^6$ data points~\cite{Boninsegna_2018, WANG2022244}.

The BISDEs method~\cite{tripura2023bayesian} partially mitigates this limitation by placing Bayesian priors over the drift and diffusion coefficients, which regularizes the estimates and reduces data requirements. However, the inference is still conditioned on bin-level statistics, so the exponential scaling with dimension is only alleviated, not eliminated.

Our framework avoids this issue entirely. The MLE step operates directly on the time series of residuals $\bs{y}$---a vector of length equal to the number of time steps---rather than on a partition of the state space. The log-likelihood (see the MLE objective in the main text and Section~\ref{app:GP}) depends only on the inner product $\bs{y}^\top \bs{K}(\bs{\Theta})^{-1} \bs{y}$ and the log-determinant $\log|\bs{K}(\bs{\Theta})|$, both of which are computed over the time axis and not over a state-space grid. As a result, the computational cost of our diffusion inference step scales only with the number of time steps and the number of candidate basis functions---both of which are independent of the state-space dimension, making the method applicable to high-dimensional systems.

\bibliographystyle{unsrt}
\bibliography{references}

\end{document}